\documentclass[10pt,twocolumn,letterpaper]{article}

\usepackage{cvpr}              % To produce the CAMERA-READY version
\usepackage{multirow}
\usepackage{tabularx}
\usepackage{siunitx}
\usepackage{bm}
\usepackage[table]{xcolor}
\usepackage{enumitem} % 导入包
\usepackage{bbding}
\usepackage{listings}
\definecolor{lightCyan}{rgb}{0.925,1,1}

\newcommand{\ours}{VINS-120K}

\newcommand{\ourBench}{VINS-4KEval}
\definecolor{cvprblue}{rgb}{0.21,0.49,0.74}
\usepackage[pagebackref,breaklinks,colorlinks,allcolors=cvprblue]{hyperref}

\def\paperID{} % *** Enter the Paper ID here
\def\confName{CVPR}
\def\confYear{2026}

\title{
VINS-120K: Ultra High-Resolution Image Editing with A Large-Scale Dataset
}

\author{
  Zhizhou Chen$^{1*}$, \hspace{0.2cm}
  Shanyan Guan$^{2*}$, \hspace{0.2cm}
  Zhanxin Gao$^1$, \hspace{0.2cm}
  En Ci$^1$, \hspace{0.2cm}
  Yanhao Ge$^2$ \\
  Wei Li$^2$, \hspace{0.2cm}
  Zhenyu Zhang$^1$, \hspace{0.2cm}
  Jian Yang$^{1}$, \hspace{0.2cm}
  Ying Tai$^{1\dagger}$ \\
  $^1$Nanjing University, China \hspace{0.2cm}
  $^2$vivo, China \hspace{0.2cm} \\
  {\tt\small 
  \{zhizhouchen,zxgao,cien\}@smail.nju.edu.cn, 
  \{zhenyuzhang,yingtai\}@nju.edu.cn
  } \\
  {\tt\small 
  \{guanshanyan,halege\}@vivo.com
  }
}
\begin{document}

\twocolumn[{%
\renewcommand\twocolumn[1][]{#1}%
\maketitle
\vspace{-1cm}
% \begin{figure*}[t]
%     \centering
%     \includegraphics[width=\linewidth]{images/teaser.pdf}
%     \caption{Comparison at 4K resolution.}
%     \label{fig:teaser}
% \end{figure*}

\begin{center}
    \centering
    \captionsetup{type=figure}
    \includegraphics[width=0.9\textwidth]{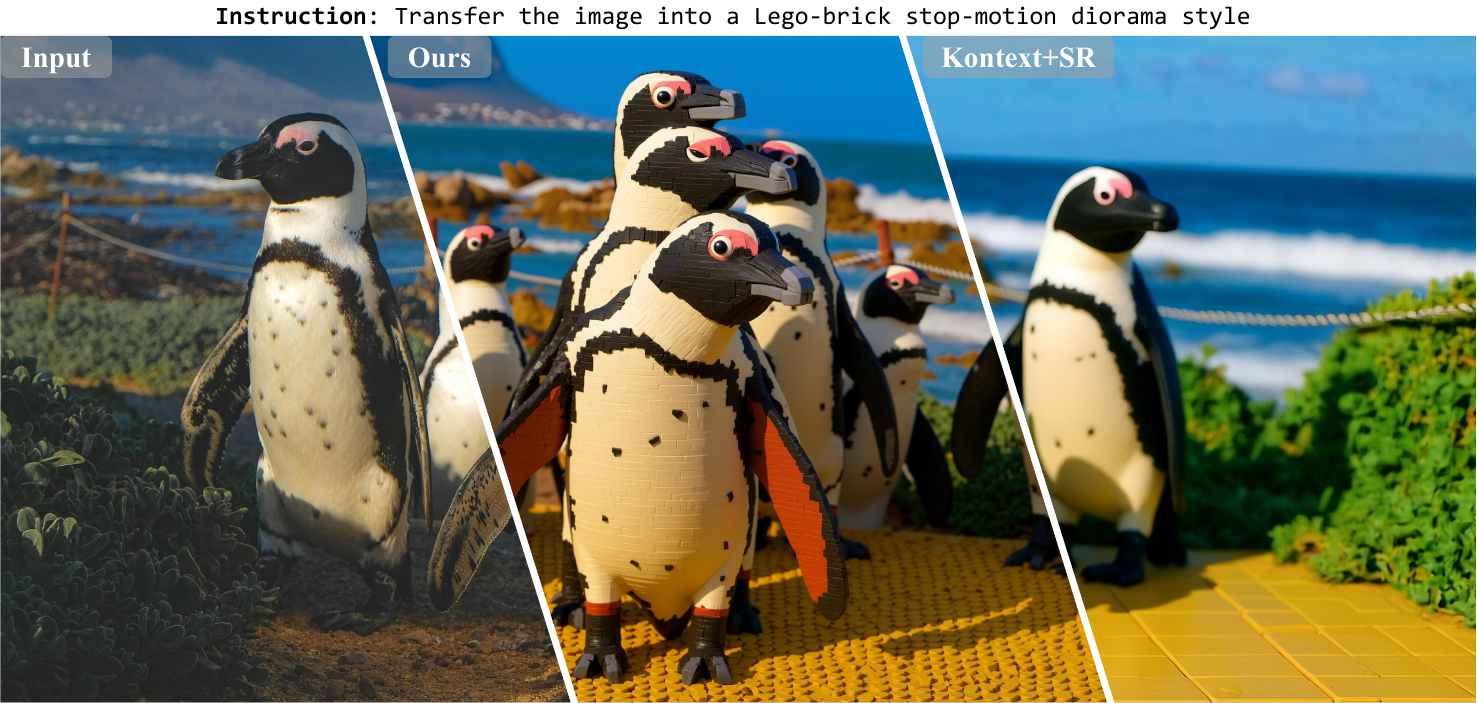}
    \vspace{-0.3cm}
\captionof{figure}{
Comparison at ultra-high-resolution editing: From left to right are the input image, our edited result, and the edited image from Kontext+SR. Kontext+SR first downsamples the input, edits it using non-high-resolution models (Kontext), and then upsamples with super-resolution techniques. Our approach outperforms by synthesizing fine-grained details and consistently adhering to instructions.
}
\vspace{-0.2cm}
\label{fig:teaser}
\end{center}

}]

\begin{abstract} 

\renewcommand{\thefootnote}{}
\footnotetext{$^*$Equal contributions. $\dagger$ indicates corresponding author.}
\renewcommand{\thefootnote}{\arabic{footnote}}

Directly editing ultra-high-resolution (UHR) images is valuable but underexplored, primarily due to the lack of high-quality data and the challenge in modeling high-frequency texture details. 
We introduce \ours{}, the first large-scale dataset for instruction-based UHR image editing, comprising 120K carefully curated triplets of instruction, input image, and edited image. 
Each image exceeds 4K resolution ($\geq$4096×4096) and is filtered through a rigorous multi-stage pipeline to ensure visual quality, instruction alignment, and aesthetic fidelity. 
Built on \ours{}, we further develop a high-frequency-aware post-adaptation strategy to extend pretrained non-high-resolution models to the UHR regime.
We also present \ourBench{}, a benchmark covering diverse editing types, to facilitate consistent evaluation in UHR settings.
Experiments confirm that our work improves fine-grained detail synthesis and texture realism in UHR image editing.
\end{abstract}

\section{Introduction}
\label{sec:intro}

Recent image editing models~\cite{instructpix2pix, emuedit, p2p, describe} have achieved strong instruction following and precise editing performance, but they are primarily designed for non-high-resolution (NHR) images, typically at 1024$\times$1024 resolution or below.
Their performance deteriorates significantly when handling ultra-high-resolution (UHR) inputs (\eg, $4096\times 4096$), manifesting as distorted images resembling noise.
Meanwhile, the growing demand for UHR content in applications such as filmmaking, advertising, and social media calls for image editing models that can perform precise edits while preserving fine-grained details at high resolution.

An intuitive solution is to downsample images, allowing existing models ~\cite{omnigen2, step1xedit, icedit, anyedit} to perform precise editing, and then upsample the images to the native resolution using advanced super-resolution models~\cite{bsrgan, realesrgan}.
% 
% While this alleviates distortion, downsampling inherently causes high-frequency information loss, and upsampling often introduces visual artifacts, leading to uncontrolled detail degradation. 
While this enables UHR editing, it inevitably introduces uncontrolled detail degradation.
Specifically, the high-frequency information lost during downsampling is difficult to faithfully recover in the subsequent upsampling stage.
As shown in Fig.~\ref{fig:teaser}, this approach (\ie,Kontext+SR) suffers from blurring and weakened instruction following.
Recent UHR datasets~\cite{diffusion4k, ultrahr, openvid} have shown the importance of fine-grained textures and semantic consistency for high-resolution visual generation.
We argue that \textit{the gap of UHR image editing lies in the absence of large-scale, high-quality UHR editing datasets and models effectively trained on such data.}

In this work, we introduce \ours{}, the first large-scale, high-quality dataset for instruction-based UHR image editing.
\ours{} contains 120K editing triplets (instruction, input image, edited image) with \textit{high fidelity} (average resolution $4656\times 4138$) and \textit{diverse instructions} (13 types across 4 categories). 
Unlike previous image editing datasets, our pairs are derived from real-world UHR videos to preserve fine-grained details.
We filter video frames using CLIP similarity and motion scores, and then generate high-quality instructions with a vision-language model through structured reasoning, global-to-local change description, and self-reflection refinement.
To improve the coverage of underrepresented edit types, we augment long-tail categories with open-source data~\cite{x2edit, nanoconsistent} that are carefully filtered and upscaled to 4K~\cite{faithdiff}.
A four-stage filtering pipeline including file checking, image quality assessment, instruction adherence filtering, and aesthetic evaluation retains only the top 20\% of samples, ensuring high data quality.
% 
% Built on \ours, we propose a High-Frequency-Aware Post-Adaptation scheme to adapt pretrained models to UHR image editing, achieving realistic details while preserving instruction adherence.
% % 
% It consists of two components: (1) Long-Token-Sequence Generalization, which rescales attention scores and the rotary base in RoPE~\cite{rope} to mitigate long-sequence degradation; and (2) Frequency-Focused Supervision, which uses a frequency-weighted loss to capture high-frequency details.
%
Built on \ours{}, we develop a simple yet effective high-frequency-aware post-adaptation strategy to address two key bottlenecks that arise when extending pretrained NHR models to the UHR regime: (i) long-sequence degradation and (ii) high-frequency details loss.
Specifically, we reformulate existing long-context modeling techniques in a resolution-aware manner and combine them with frequency-focused supervision, enabling realistic detail synthesis while preserving instruction adherence.

To evaluate UHR image editing, we present \ourBench{}, a benchmark of diverse, high-quality UHR images with varied editing instructions. Models are compared using the ImageJudge~\cite{imgedit}, VIEScore~\cite{viescore} and patch FID~\cite{ultrapixel, ultrahr} (pFID).
Applying our high-frequency-aware post-adaptation to FLUX.1-Kontext-dev~\cite{kontext}, which is pretrained at $1024\times1024$, yields strong UHR editing performance, notably reducing pFID by 28\% compared with Seedream4.0~\cite{seedream}, which primarily evaluates image details.

\section{Related Works}
\label{sec:related_works}

\paragraph{Image Editing Datasets}
The evolution of instruction-driven image editing is catalyzed by ever-larger datasets. 
While recent works like UltraEdit~\cite{ultraedit} and OmniEdit~\cite{omniedit} have scaled data volume and diversity, and X2Edit~\cite{x2edit} or ImgEdit~\cite{imgedit} have pushed resolutions to 1K-1.3K through rigorous filtering, a critical ceiling remains: no dataset supports editing beyond 1.5K resolution (Table~\ref{tab:dataset_comp}). 
Scaling to 4K is not trivial, it introduces fundamental challenges in data quality and processing due to the explosion of high-frequency details. 
Our VINS-120K addresses this gap as the first large-scale benchmark for 4K image editing.

\vspace{-11pt}
\paragraph{Text-Guided Image Editing}
Semantic editing methods aim to balance instruction fidelity with region preservation. 
Training-free approaches~\cite{p2p, pnp, masactrl} manipulate pre-trained diffusion models but often lack precision. 
Training-based methods have evolved from UNet fine-tuning (\eg, InstructPix2Pix~\cite{instructpix2pix}) to Diffusion Transformers (DiT) augmented with MLLMs~\cite{smartedit, mgie} or MoE~\cite{icedit, x2edit}, achieving strong performance on standard-resolution benchmarks ($\leq 1024 \times 1024$px).
However, directly scaling DiT-based models to 4K suffers from two limitations: (1) models trained on NHR data lack the capacity to represent UHR textures, and (2) self-attention becomes unstable on very long sequences. Our approach addresses these challenges with stabilized long-sequence attention and a frequency-focused supervision, enabling effective UHR image editing.

\begin{figure*}[t]
    \centering
\includegraphics[width=0.9\textwidth]{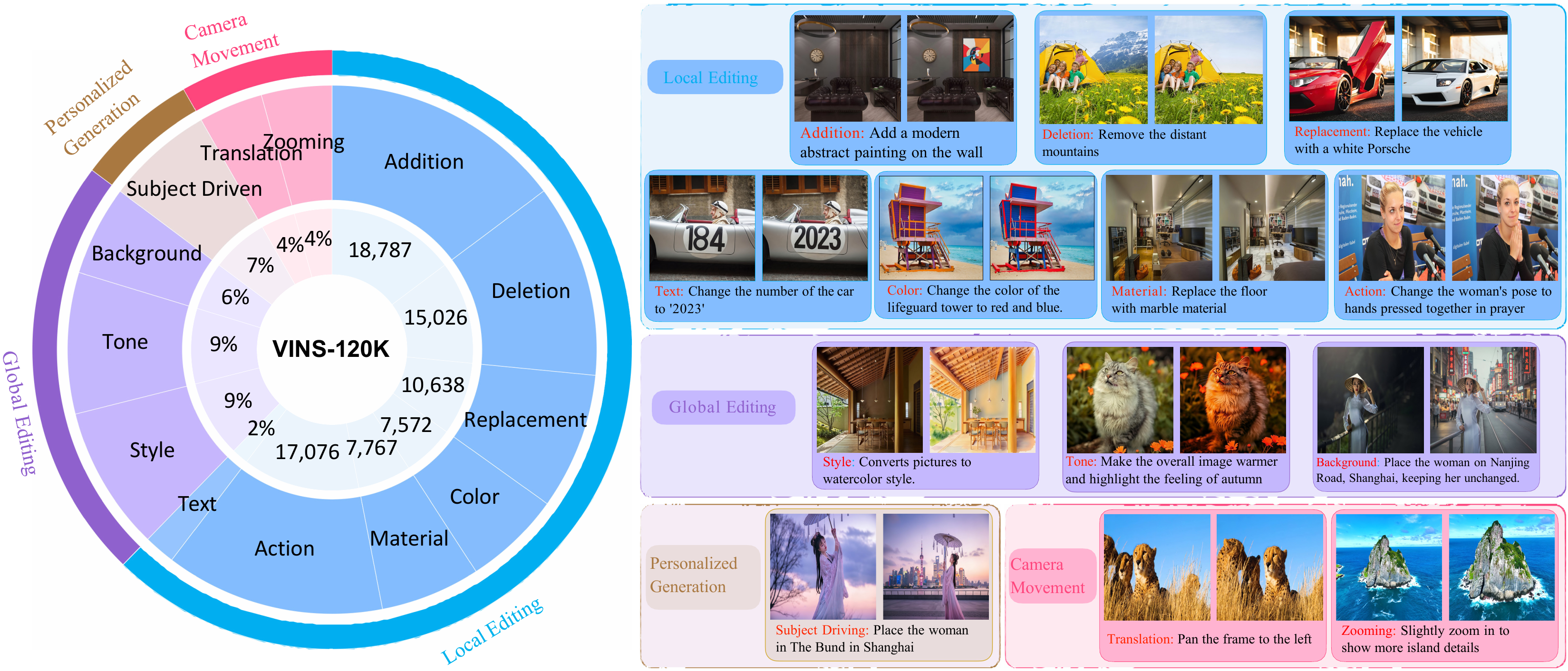}
    \vspace{-0.3cm}
    \caption{An overview and visualized examples of edited triplets (instruction, input image, edited image) for each edit type in VINS-120K.}
    \vspace{-0.6cm}
    \label{fig:edit_type}
\end{figure*}

\section{The VINS-120K Dataset}
\label{sec:VINS}

We introduce \ours{}, a large-scale, high-quality dataset addressing the lack of publicly available resources for instruction-guided UHR image editing.
\ours{} comprises 120K carefully curated UHR editing triplets (\ie, input image, edited image, and instruction), with an average resolution reaching approximately $4656\times4138$ pixels.
It covers 13 editing types grouped into four categories: local editing, global editing, camera movement, and personalized generation (see Fig.~\ref{fig:edit_type}).
We next describe the collection and annotation of the initial UHR triplets, followed by the filtering pipeline used to construct the final dataset.

\begin{figure}[t]
    \centering
    \vspace{-0.1cm}
    \includegraphics[width=0.7\linewidth]{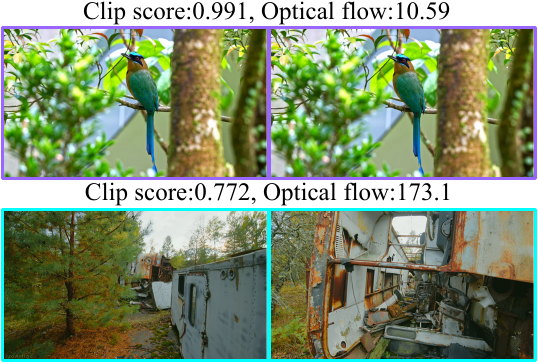}
    \vspace{-0.3cm}
    \caption{\textbf{Filtered examples of video frames.} Purple indicates similar frames (high CLIP score), while blue shows frames with semantic misalignment (high optical flow).
    }
    \vspace{-0.7cm}
    \label{fig:video_filter}
\end{figure}

    % \textbf{Data Filtering Pipeline.} We seqentially filter out broken images, low-quality images, images with inaccurate instruction, and low aesthetic images.

    % \textbf{Data Filtering Pipeline.} We seqentially filter out broken images, low-quality images, images with inaccurate instruction, and low aesthetic images.

% \begin{figure*}[t]
%     \centering
%     \includegraphics[width=\linewidth]{images/sankey.pdf}
%     \caption{
%     \textbf{Data Filtering Pipeline.} We filter images sequentially for corruption, low quality, inconsistent instructions, and poor aesthetics.
%     % \textbf{Data Filtering Pipeline.} We seqentially filter out broken images, low-quality images, images with inaccurate instruction, and low aesthetic images.
%     }
%     \label{fig:data_flow_filter}
% \end{figure*}

\subsection{Curation of UHR Editing Triplets}
\label{curation_annotation}

A central problem in building UHR editing dataset is collecting the input-edited UHR image pairs.
Beyond resolution, the UHR image should contain rich and clear texture details.
However, previous works~\cite{anyedit,imgedit} rely on fixed image-to-image pipelines, where the quality of the resulting image is inherently capped by the highest input resolution accepted by any model in the pipeline (\eg, FLUX~\cite{flux}).

Therefore, we turn to construct high-fidelity input-edited image pairs from real-world UHR videos.
Videos, as temporally continuous observations of the real world, naturally capture fine-grained visual transitions, making them an excellent source for generating realistic UHR editing pairs. 
Our construction process involves three steps. 
Given a UHR video, we first segment it into multiple clips with consistent content using PySceneDetect~\cite{pyscenedetect}. 
Next, we extract frames from each clip and combine them into candidate image pairs.
Finally, we compute semantic similarity using CLIP Score~\cite{clip} and motion score via optical flow estimation~\cite{raft} to remove pairs that are either nearly identical or exhibit excessive motion without semantic correspondence. 
As illustrated in Fig.~\ref{fig:video_filter}, this filtering effectively excludes pairs lacking meaningful transitions.

% 
% In practice, we process 8K-UHD videos ($7680\times4320$) from the UltraVideo dataset~\cite{ultravideo}, resulting in 45,000 high-quality image pairs. 
% % 
% More details are provided in the Supplementary Material.

Next, we employ Gemini-2.5-Pro~\cite{gemini} as an automatic annotator to generate editing instructions that describe visual transitions between input–edited image pairs. 
Since real-world video transitions are highly unconstrained, we guide the vision–language model (VLM) through a structured Chain-of-Thought reasoning process~\cite{cot}. 
This procedure begins with a comprehensive visual analysis of the image pair, proceeds with systematic reasoning about transformation requirements, and concludes with the formulation of precise editing instructions. 
To improve instruction quality and focus, we define a concrete action space that guides the model from global structure to local details, covering color and tone changes, camera or subject motion, and object modifications. 
In addition, to mitigate potential misjudgments and ensure instruction validity, we incorporate a self-reflection mechanism~\cite{renze2024self} that re-evaluates and refines the generated outputs through visual consistency checks. 
Details of the annotation process are provided in the Supplementary Material.

\begin{figure*}[t]
    \centering
    \includegraphics[width=0.9\textwidth]{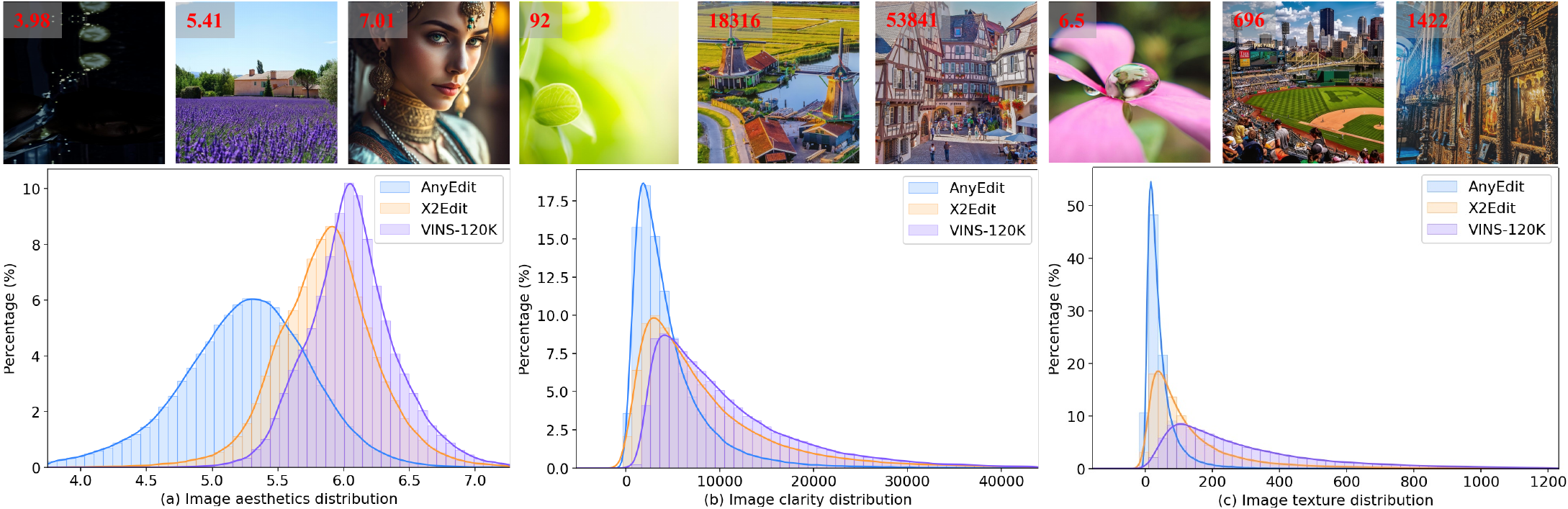}
    \vspace{-0.3cm}
    % \caption{Comparisons on image statistics between AnyEdit and X2Edit~\gsycmt{mark}}
    \caption{Comparison of image statistics between AnyEdit~\cite{anyedit} and X2Edit~\cite{x2edit} on aesthetics, clarity, and texture distributions.}
    \vspace{-0.5cm}
    \label{fig:data_comparation}
\end{figure*}

One remaining challenge is the imbalance in task distribution.
Although videos provide high-quality image pairs, certain edit types such as text modification, style change, or attribute editing are relatively rare in real-world video content. 
As noted in prior work~\cite{x2edit}, a diverse and balanced data distribution is important for model generalization.
To address this issue, we augment long-tail edit types with high-quality samples curated from recent editing datasets, including X2Edit~\cite{x2edit} and Nano-Consistent~\cite{nanoconsistent}.
We then apply the filtering pipeline described in Sec.~\ref{data_filtering} to retain samples with rich visual details, clear content, and strong aesthetic quality (see Fig.~\ref{fig:data_comparation}).
Only the samples that pass this filtering process are subsequently upscaled to 4K resolution using a modern super-resolution model~\cite{faithdiff}.
This design improves coverage of diverse edit types while mitigating the risk of learning super-resolution artifacts.
Overall, this hybrid curation strategy yields a more balanced dataset while preserving fine-grained visual details.
As shown in Tab.~\ref{tab:dataset_comp}, it also leads to clear performance gains.

% VINS-120K significantly outperforms the best values across all three metrics.

\subsection{Data Filtering}
\label{data_filtering}

\begin{figure}[t]
    \centering
    \includegraphics[width=0.9\linewidth]{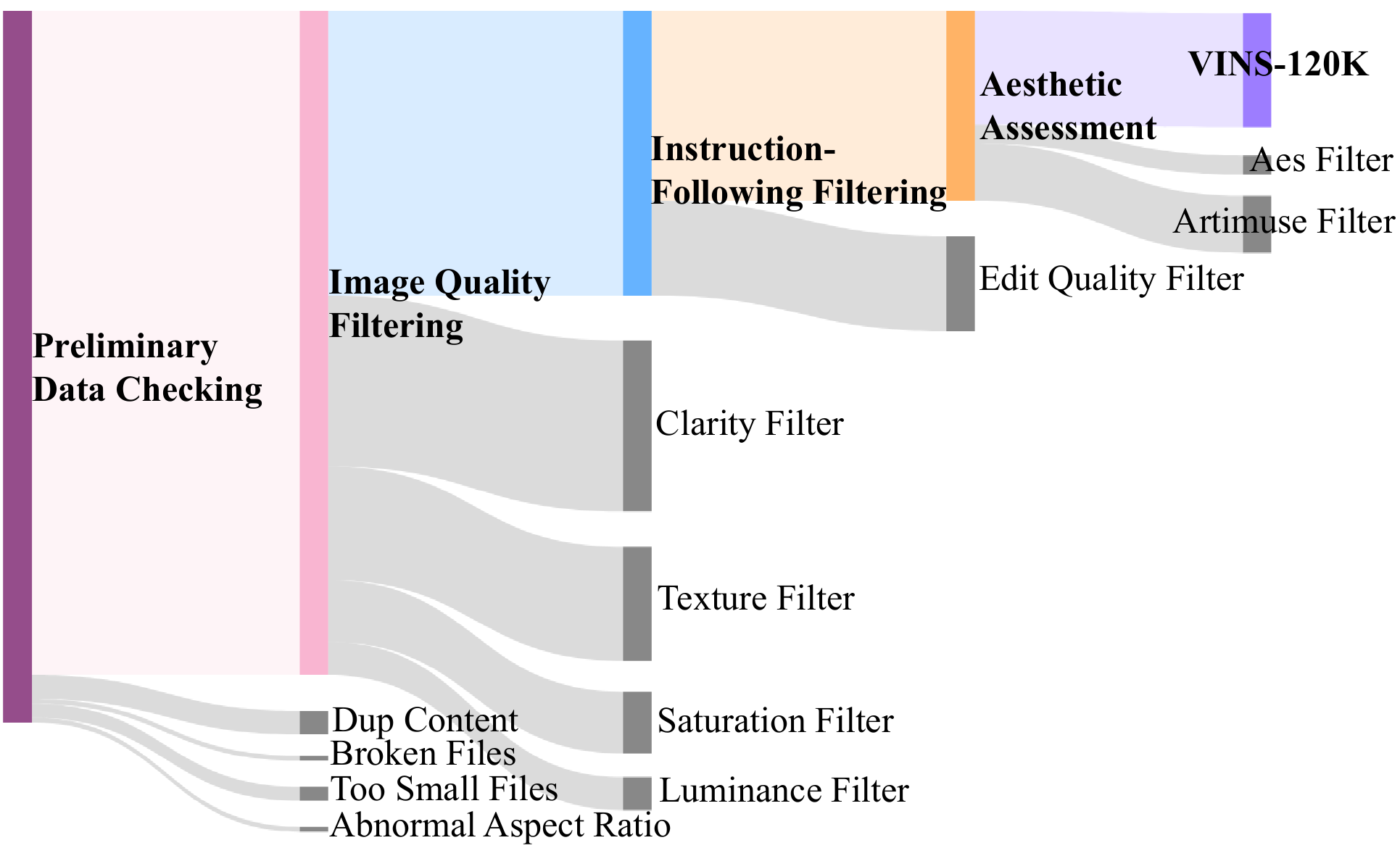}
    \vspace{-0.3cm}
    \caption{\textbf{Data Filtering Pipeline.} We filter images sequentially for corruption, low quality, inconsistent instructions, and poor aesthetics, retaining only 20\% of the highest-quality data.}
    \vspace{-0.7cm}
    \label{fig:data_flow_filter}
\end{figure}

As shown in Fig.~\ref{fig:data_flow_filter}, we propose a multi-stage filtering pipeline to refine the editing triplets curated in Sec.~\ref{curation_annotation}.
It focuses on image quality, consistency between instructions and visual changes, and aesthetic appeal. 
We describe each stage in detail below.

\vspace{-12pt}
\paragraph{Preliminary Data Checking.}
We begin with basic file validation.
Specifically, we remove corrupted or unreadable images, filter out files with abnormally small sizes, eliminate duplicates using MD5 checksums, and discard images with extreme aspect ratios to maintain a consistent spatial layout across the dataset.

\vspace{-7pt}
\paragraph{Image Quality Filtering}
In this stage, we filter high-resolution images exhibiting low visual quality through multi-dimensional filters:
\begin{itemize}[leftmargin=*]
    \item \textit{Structural Clarity}: We compute the Tenengrad gradient magnitude~\cite{focusing} to measure edge sharpness and remove samples with low gradient responses, which typically indicate blur or focus loss.
    \item \textit{Exposure Balance}: We compute the mean luminance and discard overexposed or underexposed samples that fall outside a predefined dynamic range.
    \item \textit{Color Realism}: We convert images to HSV space and remove samples with abnormal saturation statistics to exclude unrealistic color distributions.
    \item \textit{Texture Richness}: We compute Gray-Level Co-occurrence Matrix (GLCM) scores~\cite{glcm} to quantify texture diversity and remove images with atypical texture patterns or insufficient local variation.  
\end{itemize}

\vspace{-15pt}
\paragraph{Instruction-Following Filtering.} 
Consistency between editing instructions and visual changes is critical for training instruction-following models.
Although VLMs are often used to assess image editing quality, recent studies suggest that their reliability can be limited~\cite{hqedit, bpm}.
Rather than relying solely on VLMs for full evaluation, we propose a cascaded scheme.
First, we use VLMs to parse the input image and editing instruction, identifying the source and target objects involved in the modification.
Then, we apply object detection and segmentation tools to localize these objects and produce masks that decompose the image pair into edited and non-edited regions.
Within the edited regions, we compute CLIP similarity to assess instruction adherence, while $L_2$ distance in non-edited regions measures content preservation. 
By jointly considering these two metrics, we remove samples with inconsistent or irrelevant edits, yielding a refined subset in which the instructions are faithfully reflected in the visual changes.

\vspace{-15pt}
\paragraph{Aesthetic Assessment.}
At this stage, we further filter the data based on aesthetic quality.
Since aesthetic preference is inherently subjective, we leverage two complementary aesthetic assessment models—the LAION Aesthetic Predictor~\cite{laion} and Artimuse~\cite{artimuse}, a VLM specifically trained for aesthetic understanding—to jointly identify images with higher aesthetic appeal. 
This step yields a curated subset that better balances perceptual realism and aesthetic quality, bringing the dataset closer to human visual preferences.

\begin{table}[t]
\centering
\caption{Comparison of existing image editing datasets: VINS-120K is the first large-scale UHR image editing dataset and achieves the highest quality (ImageJudge-Avg).}
\vspace{-0.2cm}
\label{tab:dataset_comp}
\resizebox{\linewidth}{!}{
\begin{tabular}{l|cccc|c}
\toprule
\text{Dataset} & \text{Size} & \text{Types} & \text{Width} & \text{Height} & \text{ImageJudge-Avg} \\
\midrule
HQ-Edit~\cite{hqedit}         & 197K  & 6  & 874 & 1006   & 4.04   \\
UltraEdit~\cite{ultraedit}       & 4M    & 9  & 512 & 512 & 3.73  \\
SEED-Data~\cite{seeddataedit}  & 3.7M  & 6  & 768 & 768         & 3.58  \\
AnyEdit~\cite{anyedit}         & 2.5M  & 25 & 512 & 512 & 3.89  \\
OmniEdit~\cite{omniedit}       & 5.2M  & 7  & 1374 & 982   & 4.19  \\
ImgEdit~\cite{imgedit}         & 1.2M  & 13 & 1800 & 1200  &  4.35  \\
X2Edit~\cite{x2edit}          & 3.7M  & 14 & 1096 & 1088   & 3.98  \\
\midrule
% \rowcolor[HTML]{E6F3FF} % 浅蓝色
% \rowcolor[HTML]{E0E0E0} % 浅蓝色
\rowcolor{lightCyan}
VINS-120K       & 120K  & 13 & 4656 & 4138  & \textbf{4.45}  \\
\bottomrule
\end{tabular}
}
\vspace{-0.5cm}
\end{table}

\subsection{Dataset Statistics}

As summarized in Tab.~\ref{tab:dataset_comp}, \ours{} contains 120K high-quality image editing pairs covering 13 edit types. 
Specifically, we process 8K-UHD videos ($7680\times4320$) from UltraVideo~\cite{ultravideo} to obtain 25K high-quality triplets, and further process and filter samples from X2Edit~\cite{x2edit} and Nano-Consistent~\cite{nanoconsistent} to obtain 95K additional triplets.
As a result, \ours{} has an average image size of $4656 \times 4138$ pixels, substantially higher than existing image-editing datasets.
For quantitative comparison, we evaluate 1,000 random samples from each dataset using ImgJudge~\cite{imgedit}, as reported in Column 6 of Tab.~\ref{tab:dataset_comp}.
The results show that \ours{} achieves stronger overall image quality and instruction alignment than prior datasets.
More details of \ours{} are provided in the supplementary material.

\section{High-Frequency-Aware Post Adaptation}
\label{sec:methodlogy}

Pretrained instruction-based image editing models exhibit strong generalization to diverse editing instructions. 
However, when extended to UHR image editing scenarios, they typically struggle to accurately follow instructions and to synthesize visually realistic edits.
In this work, based on the proposed \ours{}, we aim to improve instruction adherence and enable the model to generate high-quality UHR edits through post model adaptation.
This adaptation addresses two main challenges: 
\begin{itemize}[leftmargin=*]
    \item Scaling the pretrained model to UHR resolution, which involves significantly longer token sequences compared with non-high-resolution inputs.
    \item Guiding the denoising model to better capture high-frequency image details, which is crucial for preserving fine-grained content in UHR images.
\end{itemize}

\subsection{
Long-Token-Sequence Generalization
}
\label{sec:longtoken}

The most notable change when extending to UHR image editing is the increase in token length, which poses impacts to the attention~\cite{attention} and RoPE~\cite{rope}.

\vspace{-9pt}
\paragraph{Revisiting to Attention and RoPE.}
Let $\boldsymbol{X} = \{\boldsymbol{x}_{i}\}_{i=1}^{N}$ be a sequence of token embedding, where $N$ is the token length, $\boldsymbol{x}_i \in \mathbb{R}^{d}$ is a $d$-dimensional token embedding. 
The attention layer first encoding $\boldsymbol{X}$ incorporating with position information to queries, keys representations:
\begin{equation}
    \boldsymbol{q}_{m} = f_{q}(\boldsymbol{x}_{m}, m), \qquad \boldsymbol{k}_{n} = f_{k}(\boldsymbol{x}_{n}, n),  
    % \boldsymbol{v}_{n} = f_{v}(\boldsymbol{x}_{n}, n),
\end{equation}
where $f_{*}$ are the encoding functions, and $m$,$n$ are the token positions. 
Next, the attention weight $w_{m,n}$ is calculated by
\begin{equation}
w_{m,n} = \frac{\exp\left( \frac{\boldsymbol{q}_{m}^{T}\boldsymbol{k}_{n}}{\sqrt{d}} \right)}{\sum_{j=1}^{N} \exp\left( \frac{\boldsymbol{q}_{m}^{T}\boldsymbol{k}_{j}}{\sqrt{d}} \right)}.
\label{eq:attention_score}
\end{equation}
Modern models~\cite{kontext, qwen, ragd, textcrafter} typically use RoPE to inject relative position information into the attention. 
Assume $d$ is even, and identifying $\boldsymbol{x}_i \in \mathbb{R}^{d}$ as complex vectors $\{\boldsymbol{z}_{i}\}_{i=1}^{d/2}$, where $\boldsymbol{z}_{i} = \boldsymbol{x}_{2i-1} + \boldsymbol{i} \cdot \boldsymbol{x}_{2i}$. 
RoPE applies a rotation to $\boldsymbol{z}_{i}$ based on its position $p \in \{1,2,\dots,N\}$:
\begin{equation}
    \boldsymbol{z}_{i}^{p} = \boldsymbol{z}_{i} \cdot e^{\boldsymbol{i}\boldsymbol{\theta}_{i}p}, \qquad \boldsymbol{\theta}_{i} = b^{-2i/d},
    \label{eq:rope}
\end{equation}
where $b$ is the frequency base, $\boldsymbol{\theta}_{i}p$ is the rotation angle, and the rotation is applied before encoding keys and queries. 

% where $p \in \{1,2,\dots,N\}$ is the token position, $b$ is the rotary base, and the RoPE is only applied before encoding keys and queries. 

\vspace{-10pt}
\paragraph{Rescaling Attention Score.}
Refer to Eq.~\ref{eq:attention_score}, longer token sequences tend to smooth the attention distribution, thereby weakening the discriminative responses of feature representations.
This phenomenon, often referred to as entropy shift in prior works~\cite{bigbird}, reflects the loss of attention sharpness as sequence length increases.
To counteract this effect, inspired by~\cite{du2024max, cinescale}, we introduce a resolution-aware temperature parameter $\tau>1$ to rescale the attention score:
\begin{equation}
    w^{\prime}_{m,n} = \frac{\exp\left(\tau \cdot \frac{\boldsymbol{q}_{m}^{T}\boldsymbol{k}_{n}}{\sqrt{d}} \right)}{\sum_{j=1}^{N} \exp\left( \tau \cdot \frac{\boldsymbol{q}_{m}^{T}\boldsymbol{k}_{j}}{\sqrt{d}} \right)},
\end{equation}
where $\tau$ is set to $\log\sqrt{\frac{N_{\text{UHR}}}{N_{\text{NHR}}}}$ in practice, and $N_{\text{UHR}}$ and $N_{\text{NHR}}$ denote the token sequence lengths at UHR and the pretrained model’s native resolution, respectively.

\vspace{-0.3cm}
\paragraph{Rescaling RoPE.}
Previous studies~\cite{yarn} have shown that RoPE struggles to generalize zero-shot to sequences longer than those seen during pretraining.
This issue arises because as the sequence length increases, RoPE generates rotation angles outside the range encountered during training.
The model fails to extrapolate effectively to these unseen angles, resulting in performance degradation on long sequences.
To address this, we adopt the principle from NTK-aware scaled RoPE~\cite{rope} and adapt it to the image domain.
Specifically, we rescale the rotary base $b$ to effectively interpolate the positional encodings for longer sequences: $b^{\prime}=b \cdot \sqrt{\frac{N_\text{UHR}}{N_\text{NHR}}}$.
The scaling factor $\sqrt{\frac{N_\text{UHR}}{N_\text{NHR}}}$ effectively ``compresses'' the rotation angles of the longer sequence into the native range, thereby preserving positional discriminability. % for extended contexts.

\begin{table*}[t]
\centering
\caption{
Quantitative comparison on VINS-4KEval. 
Higher values indicate better performance ($\uparrow$), except $\mathrm{FID}_{\text{patch}}$ ($\downarrow$).
}
\vspace{-0.2cm}
\label{tab:vins_4keval}
\begin{tabular}{l|*{3}{c}|*{3}{c}|c}
\toprule
\multirow{2}{*}{\textbf{Methods}} 
& \multicolumn{3}{c|}{\textbf{Image Judge}} 
& \multicolumn{3}{c|}{\textbf{VIEScore}} 
& \multirow{2}{*}{$\mathrm{pFID}\downarrow$} \\ 
\cmidrule(lr){2-4} \cmidrule(lr){5-7}
& Instr.~Adh.$\uparrow$ 
& Edit.~Qual.$\uparrow$ 
& Detail~Pres.$\uparrow$ 
& SC$\uparrow$ 
& PQ$\uparrow$ 
& Overall$\uparrow$ 
&  \\
\midrule
Seedream 4.0~\cite{seedream}     & 4.60 & 4.79 & 4.70 & 7.95 & 8.12 & 8.03 & 12.82 \\
\midrule
AnyEdit~\cite{anyedit}         & 3.24 & 3.89 & 3.57 & 4.09 & 7.32 & 5.71 & 18.44 \\
ICEdit~\cite{icedit}          & 3.76 & 4.42 & 4.09 & 5.77 & 7.81 & 6.79 & 16.69 \\
Bagel~\cite{bagel}           & 4.15 & 4.39 & 4.27 & \textbf{7.23} & 7.76 & \textbf{7.49} & 15.41 \\
Omnigen2~\cite{omnigen2}        & 4.14 & 4.54 & 4.34 & 6.73 & 7.86 & 7.29 & 18.73 \\
Step1X-Edit~\cite{step1xedit}     & 4.06 & 4.50 & 4.28 & 6.94 & 7.79 & 7.37 & 15.37 \\
\midrule
Kontext-dev~\cite{kontext}         & \underline{4.22} & \underline{4.60} & \underline{4.41} & \underline{6.95} & \underline{7.92} & 7.43 & \underline{12.66} \\
% \midrule
\rowcolor{lightCyan}
% \textbf{Ours} 
Kontext-dev + Post-Adaptation
& \textbf{4.23} & \textbf{4.70} & \textbf{4.47} & 6.89 & \textbf{7.98} & \underline{7.44} & \textbf{9.15} \\
\bottomrule
\end{tabular}
\vspace{-0.4cm}
\end{table*}

\subsection{Frequency-Focused Supervision} 

Fine-grained, high-frequency details constitute a critical source of realism in UHR images. However, standard diffusion models do not explicitly emphasize these high-frequency components and often treat them on par with low-frequency structures~\cite{dip}. To better preserve such fine-grained information, we introduce the Frequency-Focused Supervision (FFS), an auxiliary loss term designed to complement existing diffusion objectives.

Let $(\boldsymbol{x}, \boldsymbol{c}, \boldsymbol{y})$ be the triplet of input image, instruction, and target edited image, respectively.
Following the rectified flow formulation~\cite{flow}, the flow-matching loss for image editing is defined as
\begin{equation}
    \mathcal{L}_{\text{FM}}
    = \big\|\boldsymbol{\nu}(\boldsymbol{z}_t, \boldsymbol{c}, t) 
    - (\boldsymbol{\epsilon} - \boldsymbol{y}) \big\|_2^2,
\end{equation}
where $t \sim \mathcal{U}(0,1)$ is a randomly sampled continuous time step, 
$\boldsymbol{\epsilon}\!\sim\!\mathcal{N}(\boldsymbol{0},\boldsymbol{I})$ is Gaussian noise, and 
$\boldsymbol{\nu}(\cdot)$ denotes the estimated velocity field. 
The noisy latent $\boldsymbol{z}_t$ is obtained by linear interpolation: $\boldsymbol{z}_t = (1-t)\boldsymbol{x} + t\boldsymbol{\epsilon}$.
% \begin{equation}
%     \boldsymbol{z}_t = (1-t)\boldsymbol{x} + t\boldsymbol{\epsilon}.
% \end{equation}
During training, the model predicts the edited output in the latent space as $\hat{\boldsymbol{y}} = \boldsymbol{\epsilon} - \boldsymbol{\nu}(\boldsymbol{z}_t, \boldsymbol{c}, t)$.
% \begin{equation}
%     \hat{\boldsymbol{y}} = \boldsymbol{\epsilon} - 
%     \boldsymbol{\nu}(\boldsymbol{z}_t, \boldsymbol{c}, t).
% \end{equation}

We compute the frequency spectra of $\boldsymbol{y}$ and $\hat{\boldsymbol{y}}$ using orthonormal 2D discrete Fourier transforms. 
The spectral discrepancy is defined as
\begin{equation}
    \Delta \boldsymbol{F} 
    = \big|\mathrm{DFT}(\hat{\boldsymbol{y}})-\mathrm{DFT}(\boldsymbol{y})\big|
    \in \mathbb{R}^{U \times V},
\end{equation}
where $U$ and $V$ are the height and width of the frequency spectrum.
To emphasize high-frequency components, we design a dynamic frequency-weighting function:
\begin{equation}
    \mathcal{W}(\Delta\boldsymbol{F}, \alpha_t)
    = \frac{(\Delta\boldsymbol{F} + \varepsilon)^{\alpha_t}}
    {\max (\Delta\boldsymbol{F} + \varepsilon)^{\alpha_t}},
\end{equation}
where the focus intensity $\alpha_t$ is modulated according to the noise level: $\alpha_t = \alpha_{\min} + (\alpha_{\max}-\alpha_{\min})(1-t)^{\gamma}$.
% \begin{equation}
%     \alpha_t = \alpha_{\min} 
%     + (\alpha_{\max}-\alpha_{\min})(1-t)^{\gamma}.
%     \label{eq:alpha-sigma}
% \end{equation}
Here, $\alpha_{\min}$ and $\alpha_{\max}$ are the minimum and maximum values of the focus intensity, and $\gamma$ is a hyperparameter controlling how rapidly the focus intensity increases as the nose level decreases.

% applies greater weight to high frequencies and less weight to low frequencies. Additionally, we take into account the effect of the noise level: when the noise level is low ($\ie$, $t \rightarrow 0$), we assign higher weight to high-frequency components, as they are more likely to represent real details from the input image rather $x$ than noise $\epsilon$.

The resulting frequency-focused loss is defined as
\begin{equation}
    \mathcal{L}_{\text{freq}}
    = \frac{1}{UV} \sum_{u=1}^{U}\sum_{v=1}^{V}
      \mathcal{W}(\Delta\boldsymbol{F}_{uv}, \alpha_t)
      \cdot \Delta\boldsymbol{F}_{uv}.
\end{equation}
Finally, the overall training objective is $\mathcal{L} = \mathcal{L}_{\text{FM}} + \lambda\,\mathcal{L}_{\text{freq}}$, where $\lambda$ is the loss weight.
\begin{figure*}[t]
    \centering
    \includegraphics[width=\linewidth]{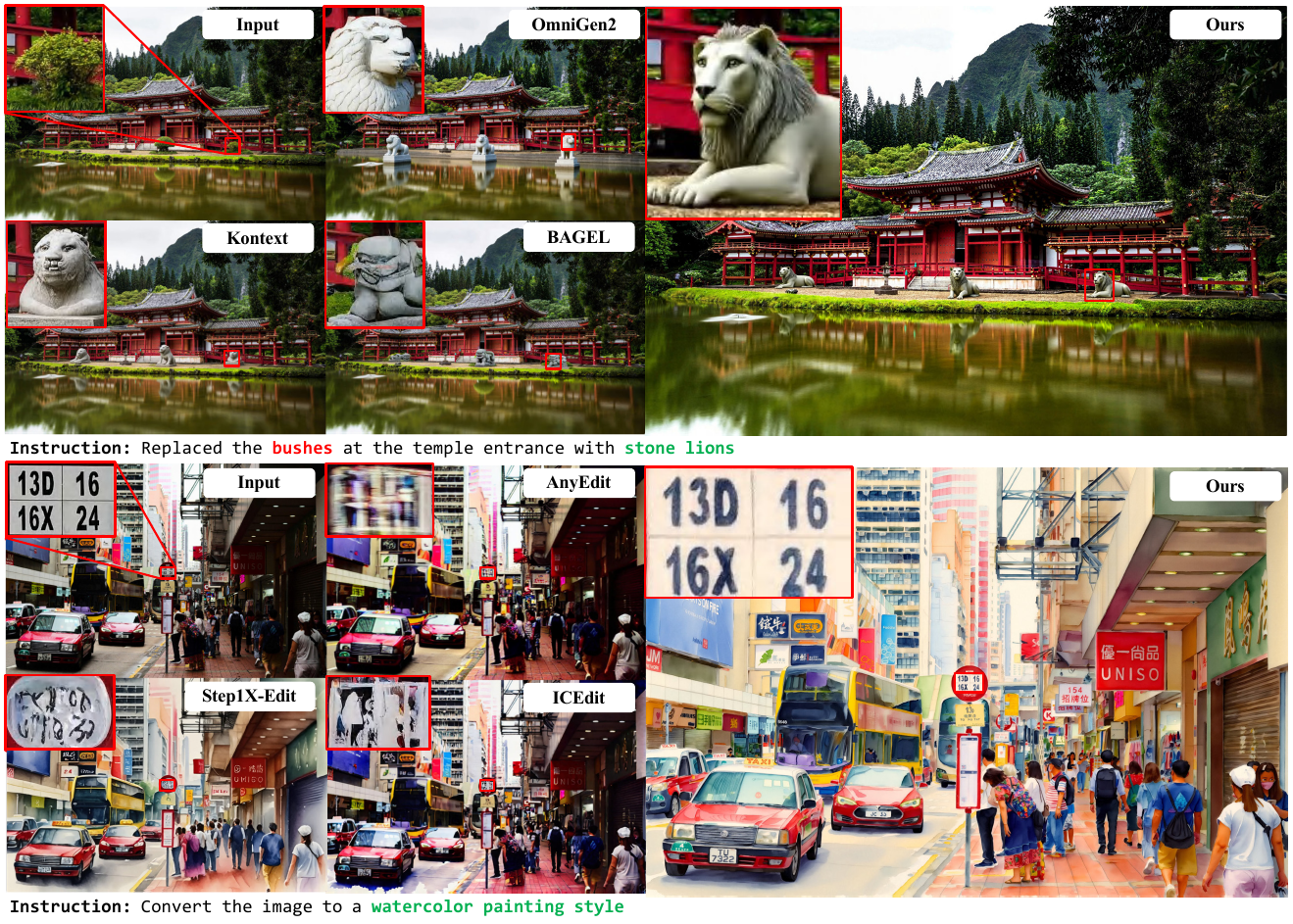}
    \vspace{-0.8cm}
    \caption{Qualitative comparisons on the VINS-4KEval benchmark.}
    \vspace{-0.6cm}
    \label{fig:qual_vins4k}
\end{figure*}

\begin{figure}[t]
    \centering
    \includegraphics[width=0.9\linewidth]{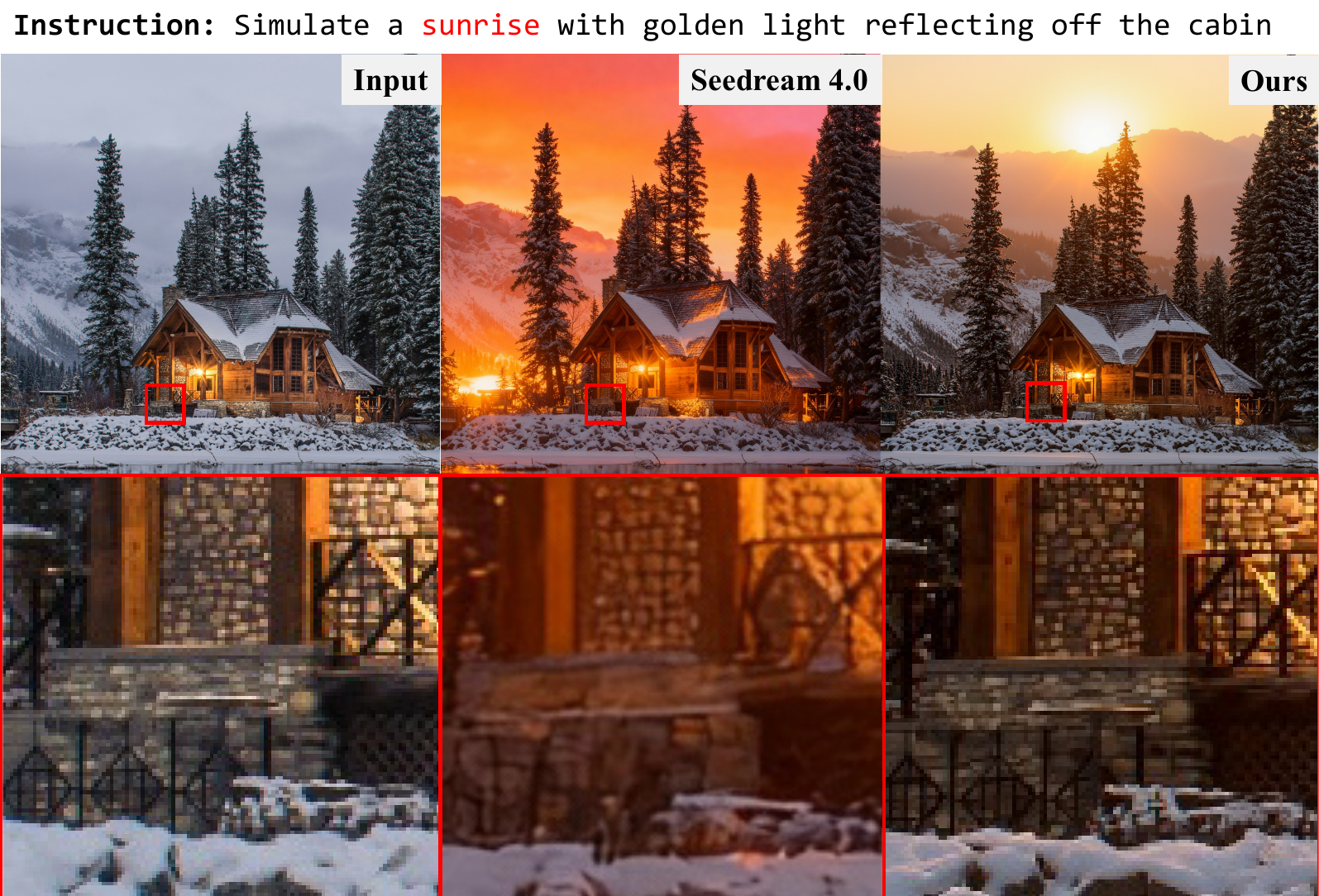}
    \vspace{-0.3cm}
    \caption{
    Qualitative comparison with Seedream 4.0.
}
    \vspace{-0.6cm}
    \label{fig:compare_w_seedream}
\end{figure}

\section{Experiment}
\label{sec:experiment}

% \subsection{Experimental Settings}

\vspace{-0.2cm}
\paragraph{Training Details.}
We use FLUX.1-Kontext-dev~\cite{kontext} as the pretrained base model and fine-tune it with LoRA~\cite{lora} of rank 32.
All training images are processed at a resolution of $4096 \times 4096$. 
We adopt the AdamW optimizer~\cite{adamw} with a learning rate of $5\times10^{-6}$. 
In the frequency-focused loss, the hyperparameters $\gamma$, $\alpha_{\min}$, $\alpha_{\max}$, and $\lambda$ are set to $2$, $0.2$, $1.2$, and $1$, respectively.

\label{sec:benchmark}
\vspace{-0.4cm}
\paragraph{Benchmark.}
To evaluate UHR image editing, we construct \ourBench{}, a 4K-resolution benchmark designed to assess both semantic accuracy and perceptual quality.
We curate high-quality Internet images from seven categories, with careful control over concept diversity, structural clarity, and texture richness.
Editing instructions are generated using our instruction generation pipeline and manually verified to remove ambiguous or unrealistic cases.
The final benchmark contains 509 UHR test samples covering 13 edit types, enabling evaluation of instruction following, structural preservation, and fine-detail quality.
Additional curation details are provided in the supplementary material.

\vspace{-0.4cm}
\paragraph{Baseline.}
We compare our method with several strong, open-source baselines, including AnyEdit~\cite{anyedit}, ICEdit~\cite{icedit}, OmniGen2~\cite{omnigen2}, Bagel~\cite{bagel}, Step1X-Edit~\cite{step1xedit}, and Kontext~\cite{kontext}. 
As most of these approaches are not designed for UHR inputs, we resize each input image to the native resolution supported by the corresponding model to ensure optimal performance.
To ensure a fair comparison, we upsample all baseline outputs back to UHR resolution using the same super-resolution model employed during the construction of the training data for \ours{}.
This design helps reduce resolution bias, so that performance differences primarily reflect editing capability.
We also report results on Seedream~4.0~\cite{seedream}, a commercial model capable of native 4K editing, as a practical reference to a strong closed-source system.
Our main reproducible comparisons remain the open-source baselines listed above.

\vspace{-0.4cm}
\paragraph{Metrics.}
We follow two widely adopted evaluation protocols, ImageJudge~\cite{imgedit} and VIEScore~\cite{viescore}, to assess the editing response accuracy. 
ImageJudge uses a finetuned VLM to evaluate edited results along three dimensions: instruction adherence, editing quality, and detail preservation.
VIEScore uses an open-sourced VLM (Qwen2.5-VL-72B~\cite{qwen2.5vl} in our implementation) to compute three metrics: Semantic Consistency (SC), Perceptual Quality (PQ), and the overall score $\sqrt{\text{SC} \times \text{PQ}}$.
In the UHR setting, we additionally report patch-FID (pFID)~\cite{ultrapixel, hiflow, ultrahr}, which measures image fidelity on local patches, following prior works.
\subsection{Main Results}

\vspace{-3pt}
\paragraph{Qualitative Comparison.}
Fig.~\ref{fig:qual_vins4k} presents qualitative comparisons on VINS-4KEval.
Because most existing baselines are not designed for UHR inputs (AnyEdit~\cite{anyedit}, ICEdit~\cite{icedit}, OmniGen2~\cite{omnigen2}, Bagel~\cite{bagel}, Step1X-Edit~\cite{step1xedit}, and Kontext~\cite{kontext}), we adopt a two-stage downsample--edit--upsample pipeline for fair comparison.
The two examples represent local and global editing scenarios.
As shown, our adapted model produces clearer and more realistic textures by editing directly in the UHR space.
In contrast, although the two-stage pipeline allows the baseline models to follow the editing instructions, the required downsampling removes high-frequency details and degrades texture quality in the final outputs.
As an additional practical reference, we also compare our method with Seedream~4.0, a strong commercial system capable of native 4K image editing.
As shown in Fig.~\ref{fig:compare_w_seedream}, adapting Kontext-dev with \ours{} yields clearer, cleaner, and more structurally faithful details than Seedream~4.0.
This comparison further suggests the effectiveness of \ours{} for improving UHR image editing quality.

\vspace{-15pt}
\paragraph{Quantitative Comparison}
Table~\ref{tab:vins_4keval} summarizes the quantitative comparison against models that were originally trained for NHR editing, as well as Seedream~4.0.
The evaluation metrics include ImageJudge and VIEScore, which primarily assess editing ability, and patch-FID (pFID), which reflects texture fidelity and overall visual quality.
Compared with Kontext-dev, the adapted model preserves its editing capability and even achieves a higher ImageJudge score.
Meanwhile, its pFID decreases from $12.66$ to $9.15$, indicating substantially improved texture fidelity after adaptation on \ours{}.
Overall, the adapted model performs favorably against other competitive NHR editing baselines in both editing performance and image fidelity.
Compared with Seedream~4.0, the adapted Kontext-dev performs worse in terms of editing ability, likely due to differences in training data scale.
However, it achieves a significantly lower pFID, suggesting that \ours{} provides effective data for UHR image editing.

% \begin{figure}[t]
%     \centering
%     % \includegraphics[width=\linewidth]{images/ablation.png}
%     \includegraphics[width=0.9\linewidth]{images/viescore_comparison.pdf}
%     \vspace{-0.2cm}
%     \caption{Model analysis on our post adaptation sceheme. SWFR is a frequence-aware loss proposed in~\cite{ultrahr}}
%     \vspace{-0.5cm}
%     \label{fig:ablation_method}
% \end{figure}

\begin{table}[t]
\centering
\caption{Analysis of post-adaptation, data curation, and cross-backbone generality on VINS-4KEval.}
\vspace{-0.3cm}
\label{ablation_tab}

\small
\setlength{\tabcolsep}{2.3pt}
\renewcommand{\arraystretch}{0.9}

\resizebox{0.95\linewidth}{!}{%
\begin{tabular}{@{}l@{\hspace{4pt}}|c c c@{}}
\toprule
\textbf{Method} & \textbf{ImageJudge} $\uparrow$ & \textbf{VIEScore} $\uparrow$ & \textbf{pFID} $\downarrow$ \\
\midrule
\rowcolor{lightCyan}
Kontext+Post-Adaptation & 4.47 & 7.44 & 9.15 \\
\midrule
w/o Post-Adaptation & 3.98 & 5.15 & 15.01 \\
\midrule
w/o Data Curation & 4.33 & 7.29 & 13.17 \\
Only Real Frames & 4.39 & 7.30 & \textbf{8.96} \\
\midrule
Qwen+SR & 4.67 & 7.93 & 18.33 \\
\rowcolor{lightCyan}
Qwen+Post-Adaptation & \textbf{4.69} & \textbf{7.97} & 11.38 \\
\bottomrule
\end{tabular}%
}

\vspace{-0.6cm}
\end{table}

\subsection{Ablation and Generalization Analysis}
\label{sec:model_analysis}

\begin{figure}[t]
    \centering
    \includegraphics[width=0.8\linewidth]{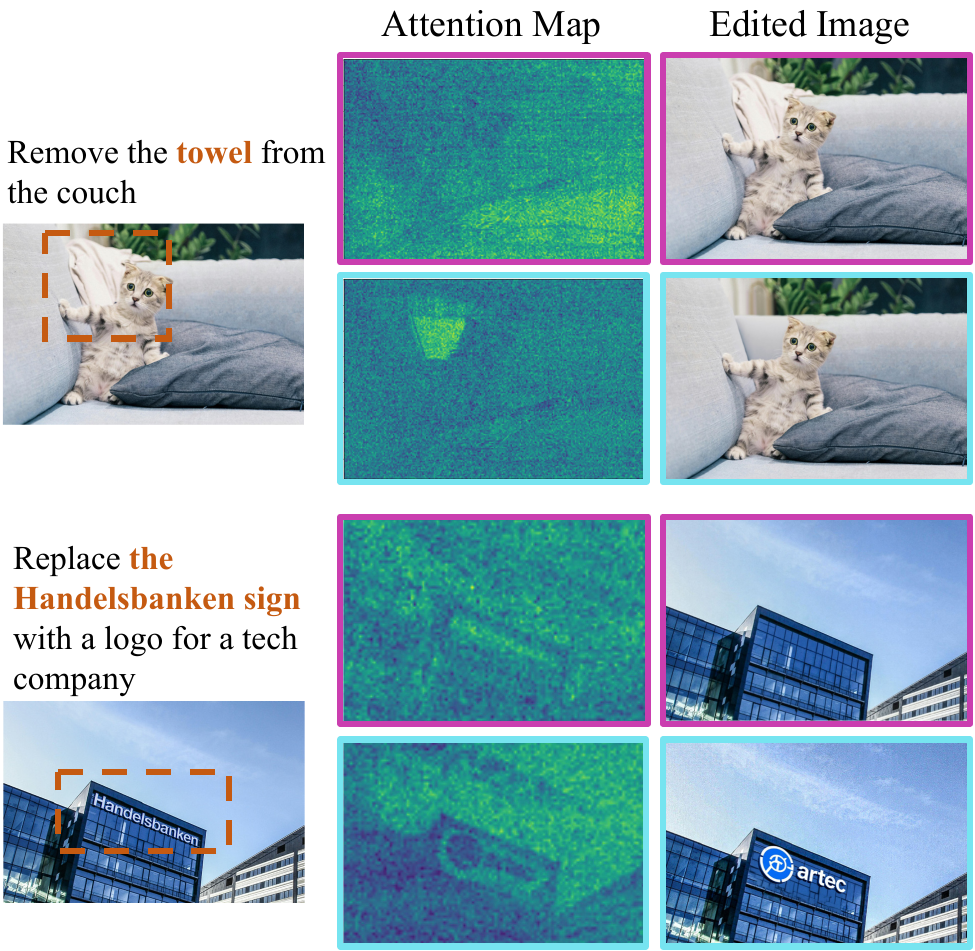}
    \vspace{-0.2cm}
    \caption{
    Ablation on attention-score rescaling. Blue: with rescaling; Purple: without rescaling. 
    }
    \vspace{-0.4cm}
    \label{fig:attn_score_scaling}
\end{figure}

\begin{figure}[t]
    \centering
    \includegraphics[width=0.9\linewidth]{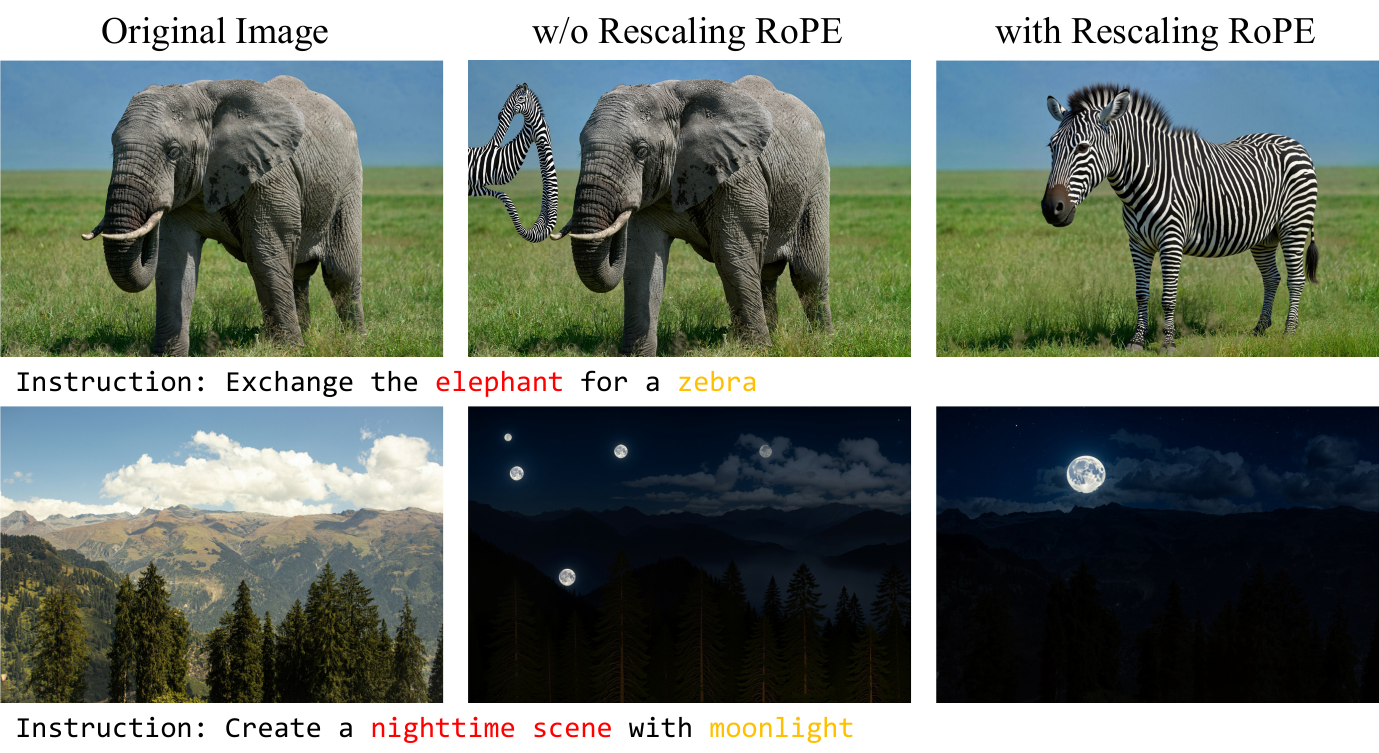}
    \vspace{-0.2cm}
    \caption{Ablation on RoPE rescaling.}
    \vspace{-0.6cm}
    \label{fig:rope_scaling}
\end{figure}

\vspace{-0.2cm}
\paragraph{Effect of Post-Adaptation.}
Table~\ref{ablation_tab} first evaluates the overall effect of the proposed post-adaptation.
From rows 1 and 2, fine-tuning Kontext-dev without methodological modifications leads to significant degradation in both editing performance and pFID.
As shown in Fig.~\ref{fig:attn_score_scaling}, attention-score rescaling allows the attention map to exhibit a more discriminative response on the target editing area.
From Fig.~\ref{fig:rope_scaling}, omitting RoPE rescaling hinders the ability to adapt to positional encodings beyond the pretraining range, resulting in semantic drift or severe local repetition.
These indicate that naive UHR scaling is insufficient and that dedicated post-adaptation is necessary for high-quality ultra-high-resolution image editing.
More detailed ablation studies are provided in the supplementary material.

\vspace{-0.4cm}
\paragraph{Effect of Data Curation.}
We further analyze the contribution of dataset construction by comparing the full setting with \textit{w/o Data Curation} and \textit{Only Real Frames}.
As shown in Tab~\ref{ablation_tab}.
From rows 1 and 3, replacing \ours{} with a UHR dataset of the same size but without careful curation slightly improves pFID, while degrading editing performance.
From rows 1 and 4, training with only native real-video frames yields superior pFID, its limited task coverage motivates curating mixed dataset, which achieves a better tradeoff: pFID from 12.66 (\textit{Kontext-dev}) to 9.15 and image editing performance from 4.39 / 7.30 (\textit{Only Real Frames}) to 4.47 / 7.44.
These results suggest that the final gains come from a balanced combination of carefully curated data and native UHR real-video pairs, rather than from scale alone.

\vspace{-0.4cm}
\paragraph{Generality Across Base Models.}
To verify that the proposed post-adaptation is not tied to a specific backbone, we apply it to QwenImage-Edit-2511~\cite{qwen} without hyperparameter retuning.
As shown in Table~\ref{ablation_tab} (rows 5–6), the adapted Qwen model achieves comparable editing ability with significantly improved detail fidelity (pFID), demonstrating that our approach generalizes beyond Kontext-dev~\cite{kontext}.
Qualitative results are provided in the supplementary material.

\vspace{-0.4cm}
\paragraph{Additional Evaluations and Ablations.} We further evaluate out-of-domain edit types and multi-turn editing scenarios.
The adapted models maintain competitive performance compared to their NHR base models, indicating that post-adaptation does not compromise general editing capability.
Details are provided in the supplementary material.

\section{Conclusion}
\label{sec:conclusion}
In this paper, we present VINS-120K, the first large-scale, high-quality dataset for instruction-based UHR image editing, consisting of over 120K triplets with fine-grained high-resolution details and diverse edit types.
To extend models pretrained on non-high-resolution images to the UHR regime, we develop a High-Frequency-Aware Post-Adaptation strategy built on \ours{}, which improves fine-detail synthesis while preserving instruction adherence.
Extensive experiments on VINS-4KEval show that the proposed dataset and adaptation strategy are effective for UHR image editing.

\vspace{-0.5cm}
\paragraph{Limitations.} UHR image editing requires substantial computational resources, limiting deployment on constrained hardware, while scaling to higher resolutions (\eg, 8K or 16K) faces challenges from exponentially longer sequences, calling for more efficient attention mechanisms and position encodings. More analysis of our limitations is provided in the supplementary material.

\section*{Acknowledgments} 

This work was supported by Natural Science Foundation of China: No. 62406135, Natural Science Foundation of Jiangsu Province: BK20241198, the Gusu Innovation and Entrepreneur Leading Talents: No. ZXL2024362.
We thank Kunlunxin for their technical support in training and evaluation on P800.

% \paragraph{Limitations.} Despite significant progress, several directions warrant future exploration. First, 4K editing requires substantial computational resources for training and inference, limiting resource-constrained deployment. More efficient architectures and quantization techniques may reduce this overhead. Second, scaling to higher resolutions (\eg, 8K or 16K) presents exponential sequence length challenges requiring innovations in attention mechanisms and position encoding. Please refer the Supplumentary material for more analysis on the limiatation.

% In this paper, we present VINS-120K, a large-scale dataset of 120K UHR image editing pairs that fills the critical resolution gap in existing datasets. Each sample is carefully curated through a hybrid paradigm combining video-derived real-world data and synthetic data. Each pair is carefully selected to ensure authentic 4K visual characteristics and precise editing instructions.
% Moreover, we introduce specialized training strategies for 4K editing, which include: (i) Noise-Aware High-Frequency Sampling (NAHFS) to enhance the model's perception of high-frequency details, and (ii) Stabilized Long-sequence Attention (SLA) to stabilize the training process for extremely long sequences. Experiments on our proposed VINS-4KEval confirm that our approach significantly enhances both editing precision and visual fidelity in ultra-high-resolution image editing.

{
    \small
    \bibliographystyle{ieeenat_fullname}
    \bibliography{main}
}

% WARNING: do not forget to delete the supplementary pages from your submission 
\clearpage
\setcounter{page}{1}
\maketitlesupplementary

\setcounter{section}{0}
\renewcommand{\thesection}{\Alph{section}}
\renewcommand{\thesubsection}{\Alph{section}.\arabic{subsection}}

\section{Overview}
\label{sec:Overview}

In this supplementary material, we present:

\begin{itemize}
\item \cref{sec:more_implement} provides a detailed description of the training and inference pipeline;
\item \cref{sec:more_evaluation} presents additional evaluations and ablation studies;
% \item \cref{sec:more_evaluation} presents additional ablation studies;
\item \cref{sec:more_dataset} shows more details about the \ours{} dataset;
\item \cref{sec:more_qualitative} includes more qualitative results using prompts from \ourBench{};
\item \cref{sec:license} outlines the license agreement required for using \ours{};
\item \cref{sec:limitation} discusses the limitations of \ours{} and potential future directions.
\end{itemize}

\vspace{-0.5cm}
\section{Implementation Details}
\label{sec:more_implement}
% 微调的LoRA模块以及采用的分布式训练框架
% 显存消耗、run time详细叙述

\subsection{Training Details}

We adopt FLUX.1-Kontext-dev\cite{kontext} as our pretrained backbone, which is a DiT-based NHR image editing model.
We fine-tune the model using LoRA\cite{lora} with rank 32, applied to a broad set of modules including the token embedder, cross-/self-attention projections (to\_q,to\_k,to\_v,to\_out), feed-forward network layers, and all corresponding blocks in both standard and single-stream transformer pathways. All training images are processed at $4096\times4096$ resolution.
Our model is trained using PyTorch with FSDP across $96\times$ NVIDIA H20 GPUs, each providing 96 GB memory.
The per-GPU batch size is 1, resulting in a global batch size of 96, with no gradient accumulation.
We use BF16 mixed precision, gradient checkpointing, and the AdamW\cite{adamw} optimizer with a learning rate of $5\times10^{-6}$. The peak memory footprint during training is approximately 95 GB per GPU.
We train for one full epoch on our proposed \ours{} dataset, containing 120K high-quality ultra high-resolution image editing pairs.
To accommodate diverse spatial resolutions, we adopt resolution-aware bucket sampling with \textit{27 predefined aspect-ratio buckets}, ensuring efficient training across heterogeneous image shapes, while no additional data augmentation is applied.
The entire training process takes approximately 11 days on 96 H20 GPUs.

\subsection{Inference Details}

Inference is performed on NVIDIA H20 GPUs using BF16 precision. \ourBench{} consists of 509 test cases, and the full evaluation at a resolution of $4096\times4096$ takes approximately 6 hours.

\section{Additional Evaluations and Ablations}
\label{sec:more_evaluation}

\subsection{Out-of-Domain Editing Evaluation}

We further evaluate whether post-adaptation to ultra-high-resolution (UHR) editing preserves the general editing capability of the pretrained models on edit types that are not explicitly covered by \ourBench{}.
We consider two out-of-domain edit types from the ImgEdit~\cite{imgedit} benchmark, namely Object Extraction and Hybrid Edit. 
To maintain consistency with the original benchmark formulation, the evaluation instructions are generated by following the sentence style of ImgEdit using the same vision-language annotation procedure described in the main paper.

\begin{figure}[t]
    \centering
    \includegraphics[width=0.95\linewidth]{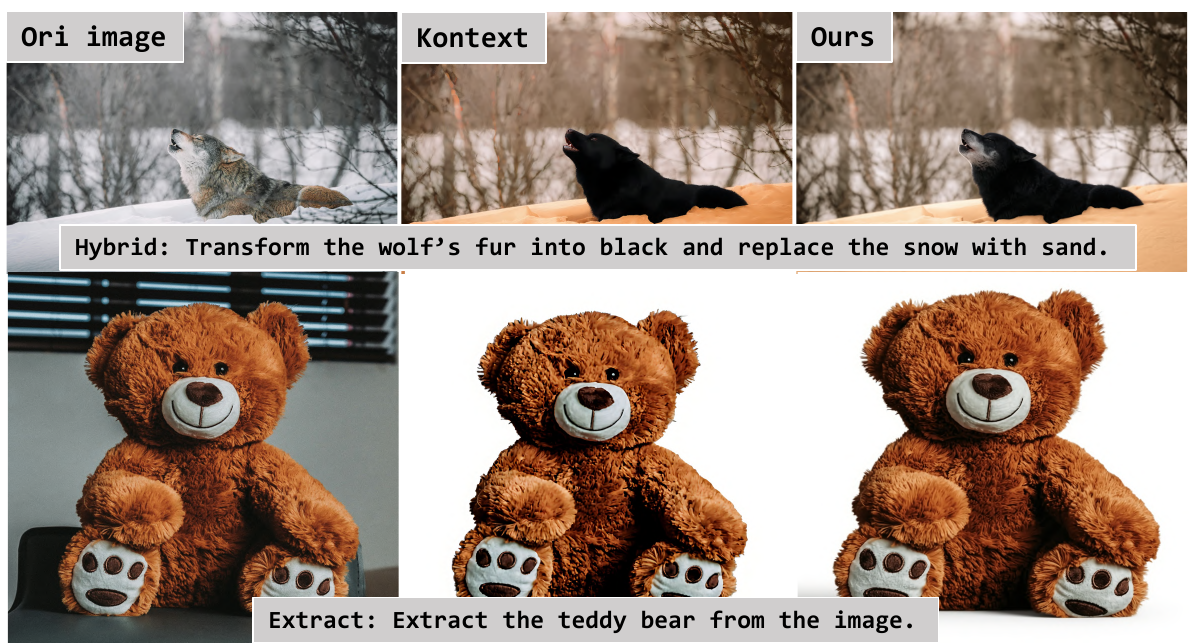}
    \vspace{-0.3cm}
    \caption{Qualitative comparisons on out-of-domain editing evaluation.}
    \vspace{-0.5cm}
    \label{fig:ood_evaluation}
\end{figure}

As shown in Tab.~\ref{rebuttal_tab}, the quantitative results show that UHR post-adaptation does not degrade out-of-domain performance. 
Instead, the adapted models maintain performance comparable to their corresponding base editors on both edit types. 
%
% In the case of the Kontext-based model, post-adaptation further improves performance on these out-of-domain instructions.
%
These observations suggest that the proposed post-adaptation mainly improves UHR detail synthesis and long-sequence stability without sacrificing the underlying instruction-following capability inherited from the pretrained editor.
Some qualitative results are shown in Fig.~\ref{fig:ood_evaluation}.

\subsection{Multi-Turn Editing Evaluation}

To assess whether the adapted model remains usable in sequential editing scenarios, we extend \ourBench{} with 20 three-turn editing samples. 
Each sample consists of a fixed input image and a sequence of three editing instructions applied successively. 
This setting is intended as an evaluation-only extension to examine whether post-adaptation to UHR editing preserves the compositional editing behavior of the original model under repeated edits.

\begin{figure*}[t]
    \centering
    \includegraphics[width=0.9\textwidth]{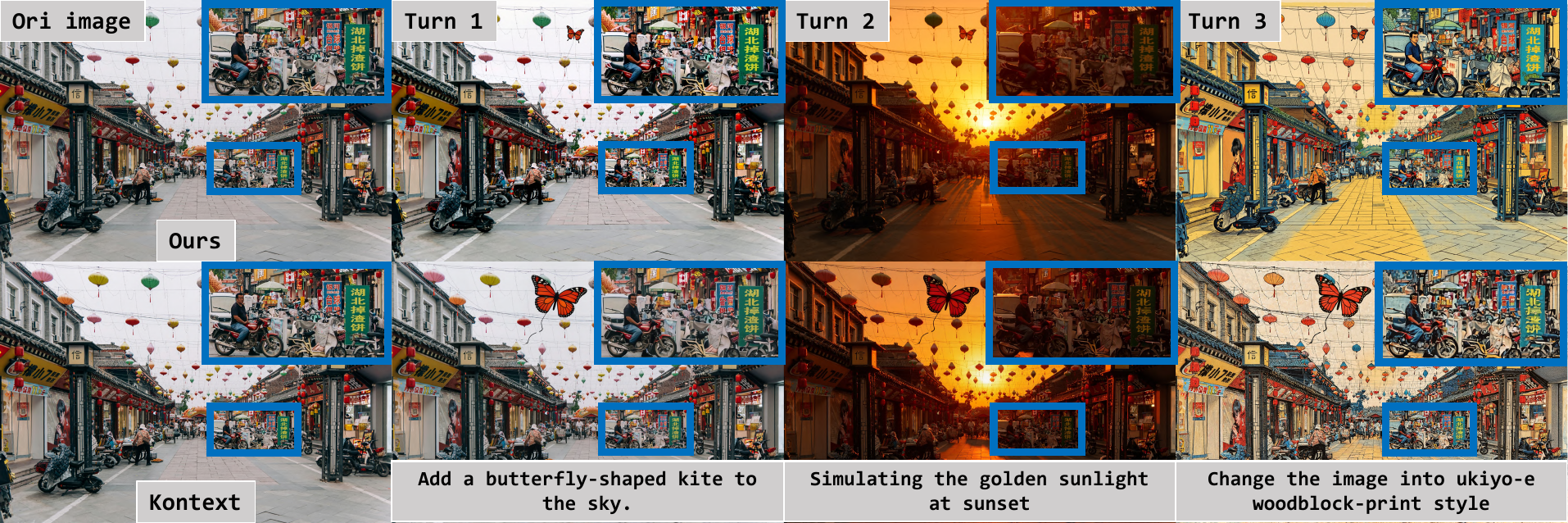}
    \vspace{-0.2cm}
    \caption{Qualitative comparison on multi-turn editing evaluation.}
    \label{fig:multi-turn_vis}
\end{figure*}

As shown in Tab.~\ref{ablation_tab}, the results indicate that the adapted models remain competitive in multi-turn editing. 
Despite being trained for UHR editing, the models preserve stable instruction following across successive turns and further improves performance on multi-turn editing relative to their base counterparts. 
This finding supports the claim that the proposed post-adaptation improves UHR fidelity while retaining general editing capability in more challenging sequential settings.
Some qualitative results are shown in Fig.~\ref{fig:multi-turn_vis}.

% \begin{table}[t]
% \centering
% \caption{Additional evaluations of out-of-domain editing and multi-turn editing. IJ: ImageJudge; VIE: VIEScore.
% }
% \vspace{-0.2cm}
% \label{rebuttal_tab}

% \footnotesize
% \setlength{\tabcolsep}{3pt}
% \renewcommand{\arraystretch}{0.92}
% \resizebox{\linewidth}{!}{%
% \begin{tabular}{l|*{2}{c}|*{2}{c}|*{2}{c}|c}
% \toprule
% \multirow{2}{*}{\textbf{Methods}} 
% & \multicolumn{2}{c|}{\textbf{Extract}} 
% & \multicolumn{2}{c|}{\textbf{Hybrid}}
% & \multicolumn{2}{c|}{\textbf{Multi-Turn}}
% & \multirow{2}{*}{pFID\(\downarrow\)} \\
% \cmidrule(lr){2-3} \cmidrule(lr){4-5} \cmidrule(lr){6-7}
% & IJ\(\uparrow\) 
% & VIE\(\uparrow\) 
% & IJ\(\uparrow\) 
% & VIE\(\uparrow\) 
% & IJ\(\uparrow\) 
% & VIE\(\uparrow\) 
% & \\
% \midrule
% w/o Post-Adaptation & 3.02 & 5.05 & 3.80 & 5.55 & 3.82 & 5.27 & 15.01 \\
% w/o Data-Curation & 4.57 & 7.55 & 4.29 & 6.93 & 4.28 & 7.31 & 13.17 \\
% \midrule
% Only Real-Frames & 4.60 & 7.60 & 4.26 & 6.96 & 4.35 & 7.42 & \textbf{9.45} \\
% \midrule
% Kontext+SR & 4.55 & 7.49 & 4.31 & 7.18 & 4.31 & 7.32 & 13.58 \\
% \rowcolor{lightCyan}
% \(\text{Ours}_{\text{Kontext}}\) & 4.62 & 7.61 & \underline{4.33} & 7.02 & 4.35 & 7.55 & \underline{10.58} \\
% Qwen+SR & \textbf{4.82} & \textbf{8.12} & 4.29 & \textbf{7.71} & \underline{4.43} & \textbf{7.94} & 18.33 \\
% \rowcolor{lightCyan}
% \(\text{Ours}_{\text{Qwen}}\) & \underline{4.72} & \underline{8.00} & \textbf{4.45} & \underline{7.70} & \textbf{4.60} & \underline{7.87} & 11.38 \\
% \bottomrule
% \end{tabular}
% }
% \vspace{-0.5cm}
% \end{table}

\begin{table}[t]
\centering
\caption{Additional evaluations of out-of-domain editing and multi-turn editing. IJ: ImageJudge; VIE: VIEScore.
}
\vspace{-0.2cm}
\label{rebuttal_tab}

\footnotesize
\setlength{\tabcolsep}{3pt}
\renewcommand{\arraystretch}{0.92}
\resizebox{\linewidth}{!}{%
\begin{tabular}{l|*{2}{c}|*{2}{c}|*{2}{c}|c}
\toprule
\multirow{2}{*}{\textbf{Methods}} 
& \multicolumn{2}{c|}{\textbf{Extract}} 
& \multicolumn{2}{c|}{\textbf{Hybrid}}
& \multicolumn{2}{c|}{\textbf{Multi-Turn}}
& \multirow{2}{*}{pFID\(\downarrow\)} \\
\cmidrule(lr){2-3} \cmidrule(lr){4-5} \cmidrule(lr){6-7}
& IJ\(\uparrow\) 
& VIE\(\uparrow\) 
& IJ\(\uparrow\) 
& VIE\(\uparrow\) 
& IJ\(\uparrow\) 
& VIE\(\uparrow\) 
& \\
\midrule
w/o Post-Adaptation & 3.02 & 5.05 & 3.80 & 5.55 & 3.82 & 5.27 & 15.01 \\
w/o Data-Curation & 4.47 & 7.55 & 4.49 & 6.93 & 4.21 & 7.31 & 13.17 \\
\midrule
Only Real-Frames & 4.50 & 7.60 & 4.47 & 6.96 & 4.27 & 7.42 & \textbf{9.45} \\
\midrule
Kontext+SR & 4.55 & 7.49 & \textbf{4.71} & 7.18 & 4.18 & 7.32 & 13.58 \\
\rowcolor{lightCyan}
\(\text{Ours}_{\text{Kontext}}\) & 4.52 & 7.61 & 4.59 & 7.02 & 4.27 & 7.55 & \underline{10.58} \\
Qwen+SR & \textbf{4.69} & \textbf{8.12} & 4.66 & \textbf{7.71} & \underline{4.34} & \textbf{7.94} & 18.33 \\
\rowcolor{lightCyan}
\(\text{Ours}_{\text{Qwen}}\) & \underline{4.55} & \underline{8.00} & \underline{4.67} & \underline{7.70} & \textbf{4.52} & \underline{7.87} & 11.38 \\
\bottomrule
\end{tabular}
}
\vspace{-0.5cm}
\end{table}

\subsection{Long-Token-Sequence Generalization}

\begin{figure}[t]
    \centering
    \includegraphics[width=0.75\linewidth]{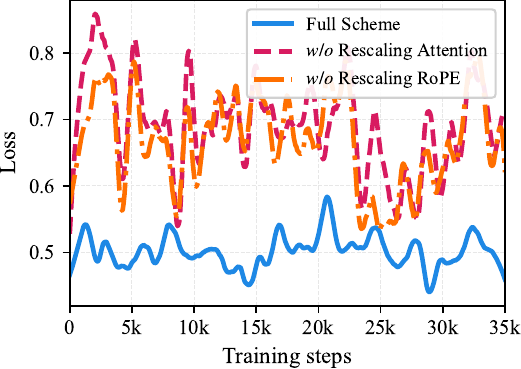}
    \vspace{-0.2cm}
    \caption{
    Loss curves of models trained without rescaling attention, without rescaling RoPE, and with Long-Token-Sequence Generalization.
    }
    \vspace{-0.2cm}
    \label{fig:training_curvers}
\end{figure}

To evaluate the effectiveness of our Long-Token-Sequence Generalization, we compare the training loss trajectories across three configurations, as shown in Fig.~\ref{fig:training_curvers}. Under the same number of training steps, the full model exhibits noticeably more stable optimization and converges to a lower loss.
As discussed in Sec.\ref{sec:model_analysis}, removing attention-score rescaling produces over-smoothed attention distributions that suppress token-level distinctions, preventing the model from generating discriminative responses in the target editing regions. This leads to inaccurate or incomplete edits. Similarly, omitting RoPE rescaling hinders the model’s ability to adapt to positional encodings beyond the pretraining range, often resulting in semantic drift or severe local repetition.
Both issues introduce noticeable instability and oscillations during training.

\subsection{Ablation of Frequency-Focused Supervision}

We report the ablation results of frequency-focused supervision on the full evaluation set in Table~\ref{tab:ablation_ffs}. The results show that introducing explicit supervision on high-frequency components consistently improves the synthesis of perceptually meaningful details in high-resolution images, leading to overall performance gains.

\begin{figure*}[t]
    \centering
    \includegraphics[width=0.85\textwidth]{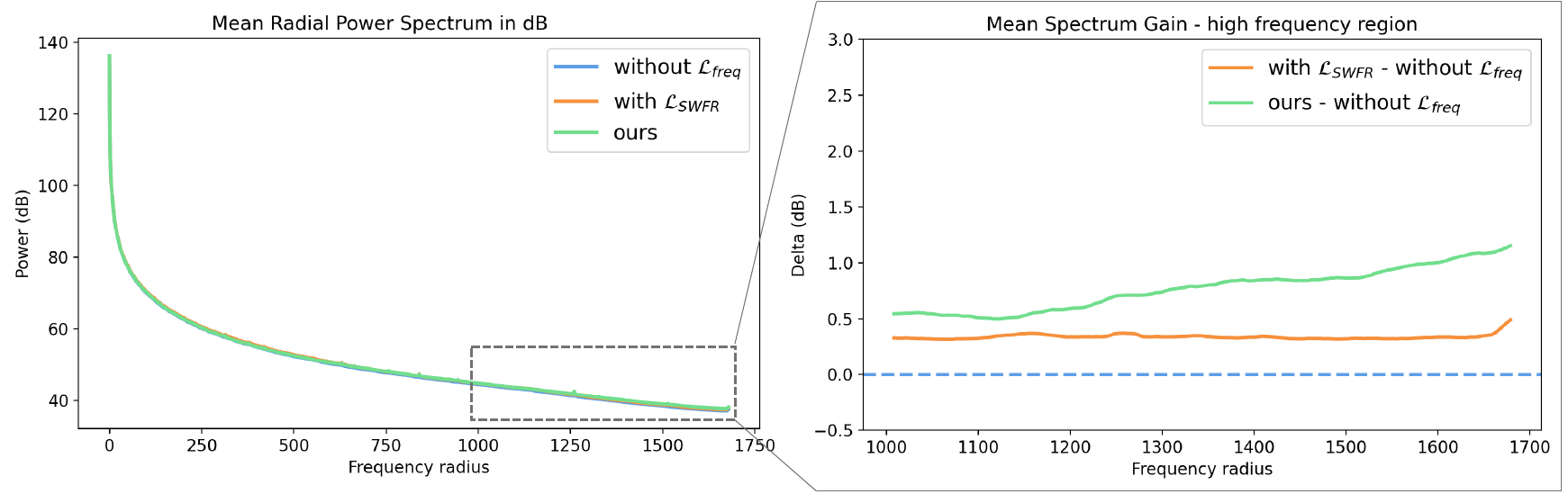}
    \vspace{-0.2cm}
    \caption{Spectral density analysis of the generated images.}
    \label{fig:spectral_density}
\end{figure*}

However, as indicated by the spectral density analysis in Fig.~\ref{fig:spectral_density}, applying strong high-frequency emphasis throughout the entire denoising process, while increasing the amount of high-frequency content, also tends to amplify noise-dominated high-frequency signals in the early stages of denoising. This effect introduces unrealistic artifacts, degrades \textbf{pFID}, and ultimately harms the realism of fine details.

In contrast, our method adaptively adjusts the strength of high-frequency supervision according to the noise level. Specifically, the supervision is attenuated under high-noise conditions and gradually strengthened as denoising proceeds. This adaptive design improves the stability of early-stage reconstruction while enabling more accurate and visually faithful synthesis of high-frequency details in later stages.

% \begin{table}[t]
% \centering
% \scriptsize
% \setlength{\tabcolsep}{5pt}   % 控制列间距，3-6pt之间可微调
% \renewcommand{\arraystretch}{1.15} % 控制行距

% \caption{Ablation results for Frequency-Focused Supervision, including the flow-matching loss $\mathcal{L}_{FM}$ and the high-frequency loss $\mathcal{L}_{SWFR}$ from UltraHR~\cite{ultrahr}.}
% \vspace{-0.2cm}
% \label{tab:ablation_ffs}
% \resizebox{\columnwidth}{!}{%
% \begin{tabular}{lccc}
% \toprule
%  $\textbf{Loss Function}$ & \textbf{ImageJudge}$\uparrow$ & \textbf{VIEScore}$\uparrow$ &
%  \textbf{pFID$\downarrow$} \\
% \midrule
% $\mathcal{L}_{FM}$ & 4.24 & 7.29 & 9.61 \\
% $\mathcal{L}_{FM}+\mathcal{L}_{SWFR}$ & 4.24 & 7.35 & 10.02 \\
% \rowcolor{lightCyan}
% $\mathcal{L}_{FM}+\mathcal{L}_{freq}$ & 4.31 & 7.47 & 9.51 \\
% \bottomrule
% \end{tabular}
% }
% \vspace{-0.5cm}
% \end{table}

\begin{table}[t]
\centering
\scriptsize
\setlength{\tabcolsep}{5pt}   % 控制列间距，3-6pt之间可微调
\renewcommand{\arraystretch}{1.15} % 控制行距

\caption{Ablation results for Frequency-Focused Supervision, including the flow-matching loss $\mathcal{L}_{FM}$ and the high-frequency loss $\mathcal{L}_{SWFR}$ from UltraHR~\cite{ultrahr}.}
\vspace{-0.2cm}
\label{tab:ablation_ffs}
\resizebox{\columnwidth}{!}{%
\begin{tabular}{lccc}
\toprule
 $\textbf{Loss Function}$ & \textbf{ImageJudge}$\uparrow$ & \textbf{VIEScore}$\uparrow$ &
 \textbf{pFID$\downarrow$} \\
\midrule
$\mathcal{L}_{FM}$ & 4.41 & 7.27 & 9.25 \\
$\mathcal{L}_{FM}+\mathcal{L}_{SWFR}$ & 4.40 & 7.32 & 9.64 \\
\rowcolor{lightCyan}
$\mathcal{L}_{FM}+\mathcal{L}_{freq}$ & 4.47 & 7.44 & 9.15 \\
\bottomrule
\end{tabular}
}
\vspace{-0.5cm}
\end{table}

\section{More Details About the \ours{} Dataset}
\label{sec:more_dataset}

\begin{figure}[t]
    \centering
    \includegraphics[width=0.9\linewidth]{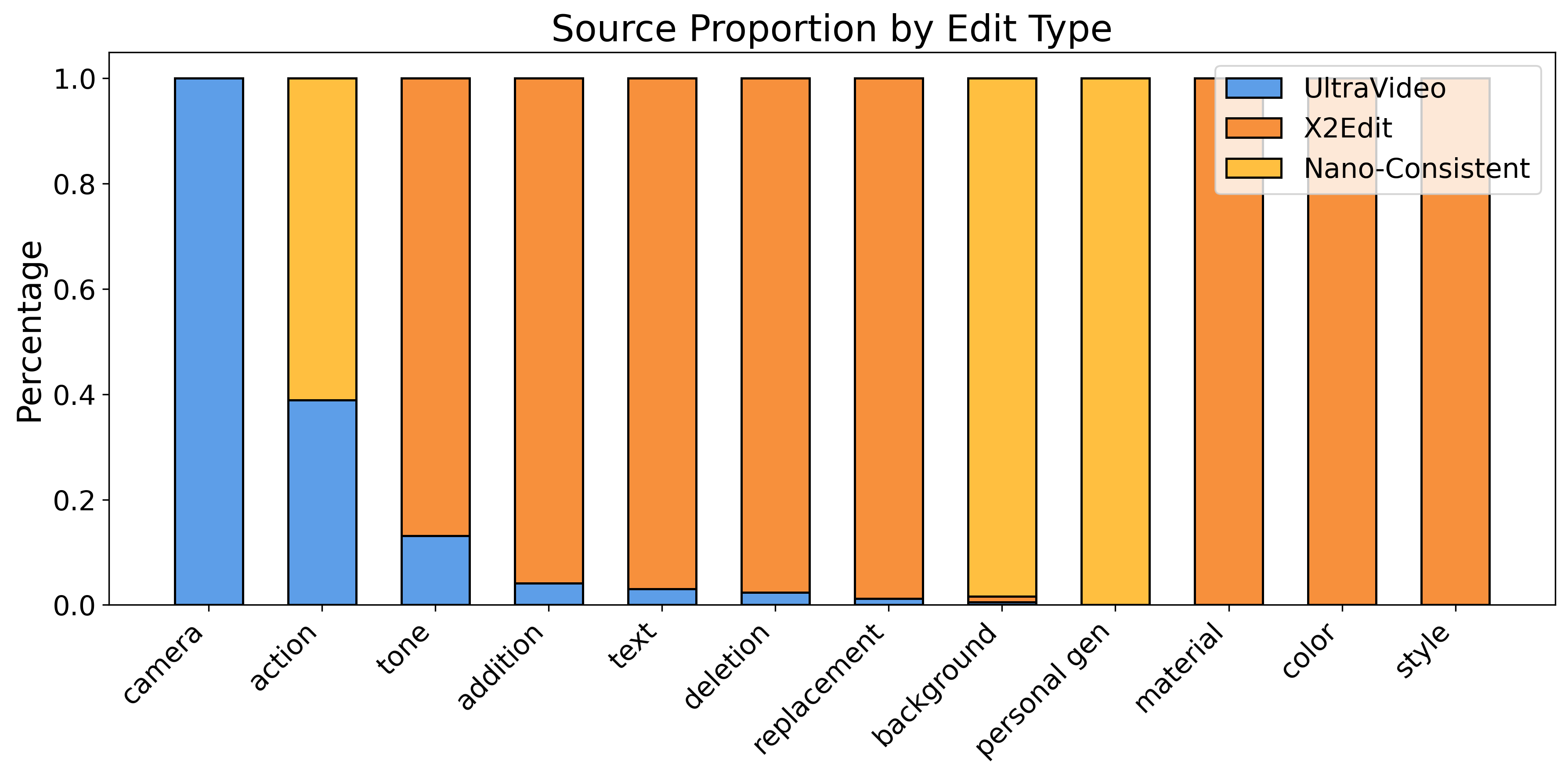}
    \vspace{-0.3cm}
    \caption{Distribution of editing types in VINS-120K across different data sources.}
    \vspace{-0.5cm}
    \label{fig:data_source_ablation}
\end{figure}

\subsection{More Statistical Information of \ours{}}

\ours{} is built from two complementary sources: 25K native UHR pairs collected from real videos and 95K curated-and-upscaled pairs derived from existing editing datasets. After filtering, we estimate the average ImageJudge scores for each source by repeatedly sampling subsets and averaging the results. The resulting scores for X2Edit, Nano-Consistent, and real videos are $4.35$, $4.39$, and $4.56$, respectively, all higher than that of X2Edit~\cite{x2edit}.

We further summarize the edit-type coverage of each source in Fig.~\ref{fig:data_source_ablation}. Real videos mainly contribute natural high-frequency transitions, while external datasets complement relatively rare edit types, indicating that the two sources play different but complementary roles in dataset construction.

\begin{figure*}[t]
    \centering
    \includegraphics[width=0.9\textwidth]{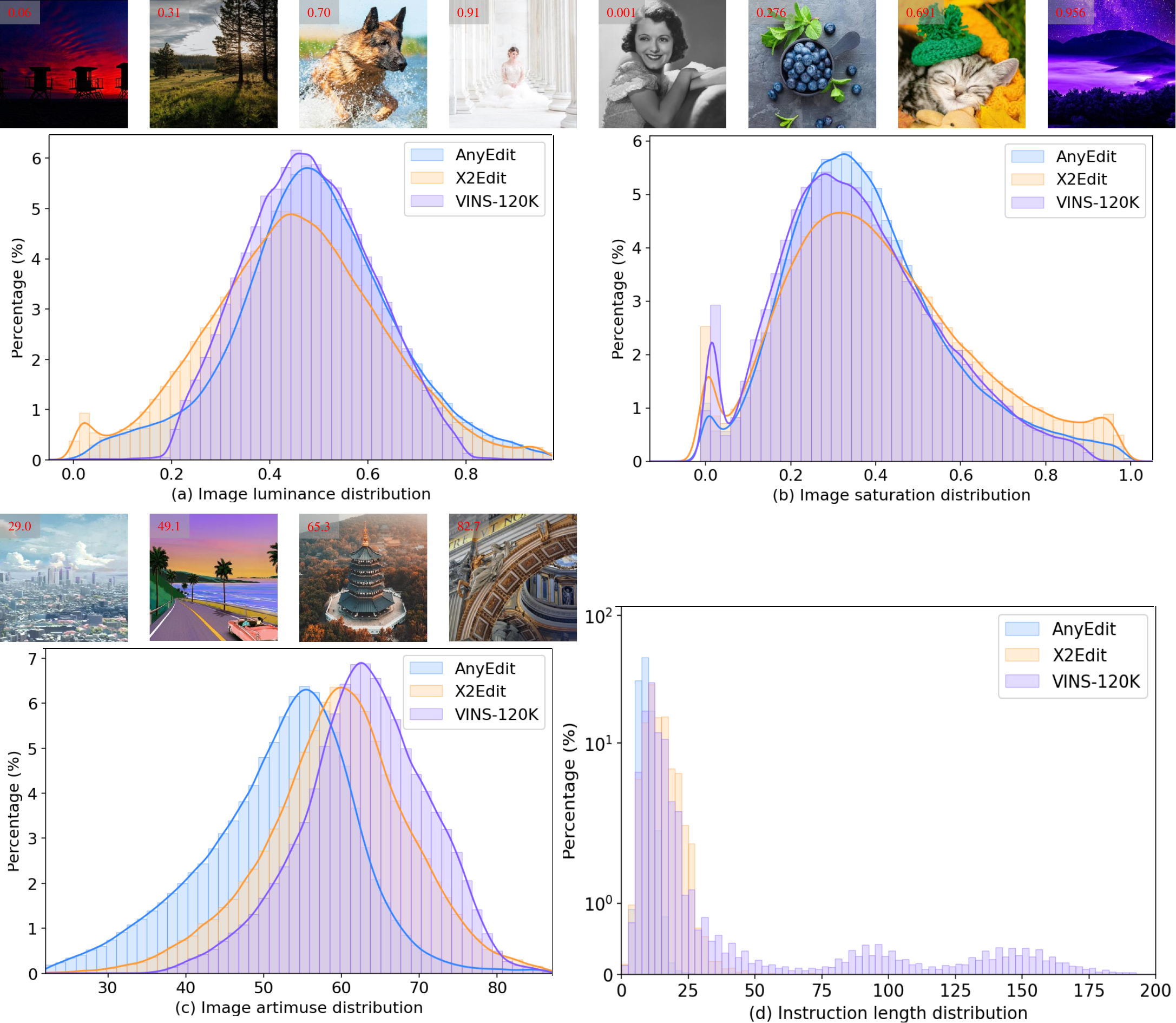}
    \vspace{-0.2cm}
    \caption{Comparison of image statistics between AnyEdit~\cite{anyedit} and X2Edit~\cite{x2edit} on luminance, saturation, and artimuse distributions.}
    \label{fig:data_comparation2}
\end{figure*}

We also compare \ours{} with AnyEdit~\cite{anyedit} and X2Edit~\cite{x2edit} in Fig.~\ref{fig:data_comparation} and Fig.~\ref{fig:data_comparation2}. The main observations are as follows:

\begin{enumerate}[leftmargin=1.5em,itemsep=2pt,topsep=2pt]
    \item \textbf{LAION aesthetic scores.}
    \ours{} shows higher LAION aesthetic scores than both AnyEdit and X2Edit, suggesting better overall visual quality under this metric.

    \item \textbf{Image sharpness.}
    Compared with AnyEdit and X2Edit, \ours{} contains fewer out-of-focus or blurry samples, reflecting the effect of our filtering process.

    \item \textbf{Texture complexity.}
    \ours{} exhibits higher texture-related statistics than AnyEdit and X2Edit, suggesting richer high-frequency details.

    \item \textbf{Brightness distribution.}
    \ours{} shows a more balanced brightness distribution around mid-range values, while AnyEdit and X2Edit include more underexposed and overexposed samples.

    \item \textbf{Color saturation.}
    \ours{} contains fewer abnormally over-saturated samples than AnyEdit and X2Edit. 
    
    \item \textbf{Artistic aesthetics.}
    \ours{} achieves higher Artimuse aesthetic scores, indicating stronger performance under this aesthetic metric.

    \item \textbf{Instruction length.}
    Editing instructions in \ours{} are generally longer than those in AnyEdit and X2Edit, suggesting that they may encode richer semantic information.
\end{enumerate}

Overall, these statistics suggest that \ours{} offers improved sample quality and broader instruction expressiveness than AnyEdit and X2Edit, while benefiting from the complementary strengths of real-video data and external editing datasets.

\subsection{Editing Instruction Annotation}

\begin{figure*}[t]
    \centering
    \includegraphics[width=0.9\textwidth]{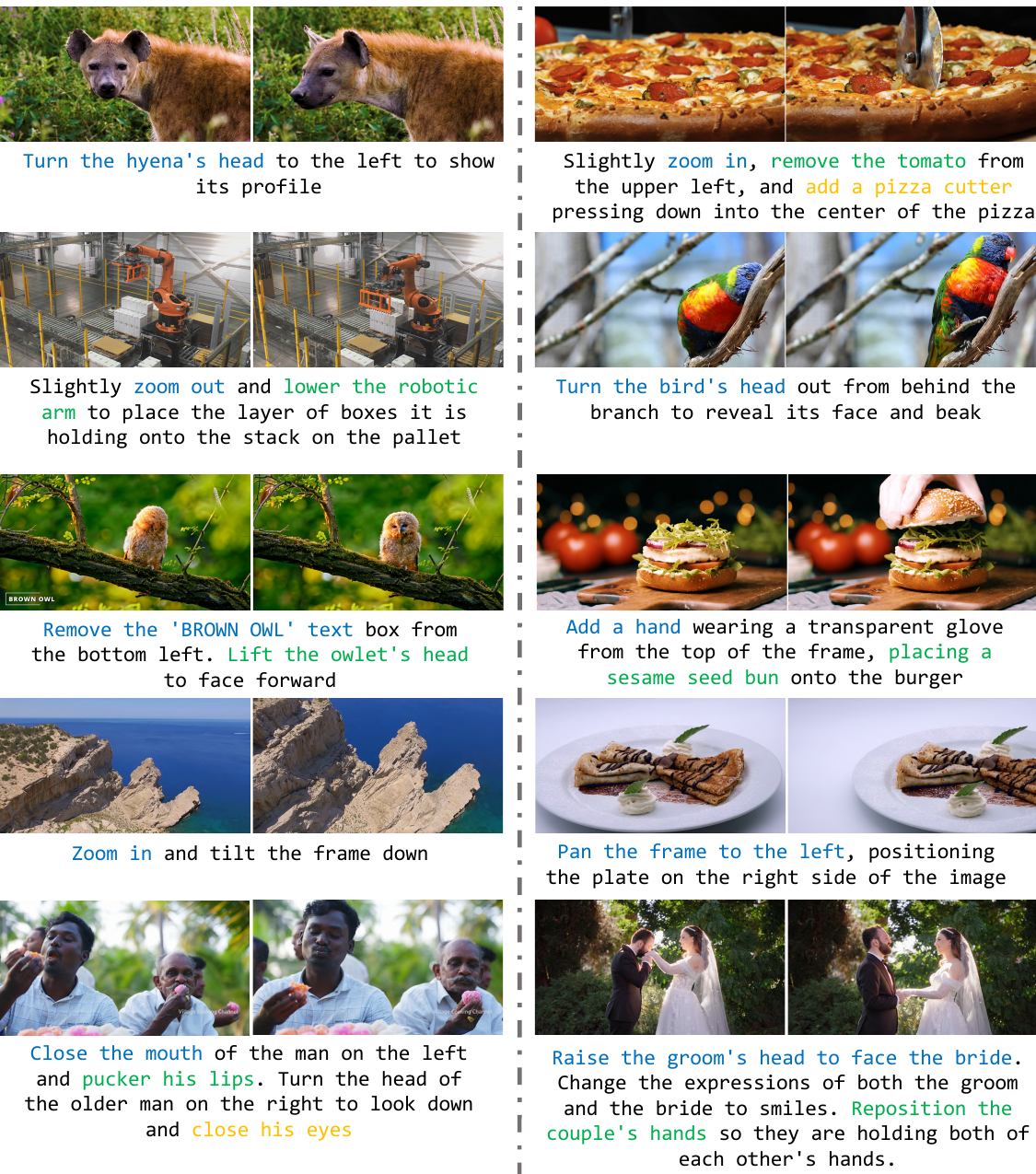}
    \vspace{-0.2cm}
    \caption{Examples of Editing Instruction Annotation performed by our pipeline.}
    \label{fig:more_video_examples}
\end{figure*}

To bridge the potential and unconstrained semantic discrepancies between video frame pairs, we adopt Gemini-2.5-Pro\cite{gemini}, a state-of-the-art vision–language model (VLM), to annotate the visual transition process. The design rationale of our prompts follows Sec.~\ref{curation_annotation}. Based on this setup, we generate editing instructions of varying lengths that are detailed, accurate, and executable. The full prompt is provided below, and Fig.~\ref{fig:more_video_examples} further illustrates example annotations demonstrating their quality.

\vspace{-0.5cm}
\paragraph{Full Prompt Used for VLM Annotation.} 

\textit{You are a ``Vision-to-Edit'' expert. Your task is to compare the first image and the second image, and output minimal, clear, and executable editing instructions to transform the first image into the second image.}

\textit{Your task must follow these requirements:}
\begin{enumerate}
    \item \textit{First, understand the content of both images and coherently describe the two frames in detail, including the environment, main subjects, their appearance, and key characteristics.}
    \item \textit{Only describe factual and visibly observable differences. Do not speculate about unseen content, nor use uncertain expressions such as ``possibly,'' ``seems,'' or ``probably.''}
    \item \textit{First consider global camera changes (translation, zoom, rotation, cropping), and then describe object-level changes such as addition, deletion, replacement, movement, size, pose, color, lighting, and text.}
    \item \textit{Editing instructions must be concise, precise, executable, written in active voice, and starting with a verb.}
    \item \textit{Do not output your reasoning process or intermediate steps. If uncertain, omit that detail.}
    \item \textit{Avoid vague pronouns or ambiguous references, even though the prompt mentions ``first image'' and ``second image.''}
    \item \textit{If the two images are almost identical, output ``NO\_CHANGE.''}
\end{enumerate}

\textit{Below are examples of atomic editing instructions:}
\begin{itemize}
    \item \textit{camera movement: pan the frame to the right; rotate the frame $20^\circ$ clockwise; slightly zoom in.}
    \item \textit{object movement: move the cup to the right.}
    \item \textit{action change: raise the person's right hand; turn the bird's head to the right.}
    \item \textit{object addition: add a car on the road.}
    \item \textit{object deletion: remove the pedestrian in the background.}
    \item \textit{object replacement: replace the bird with a butterfly.}
    \item \textit{background change: change the background to a country road.}
    \item \textit{color change: change the goose's feathers to black.}
    \item \textit{text change: change the sign text from ``neutral'' to ``health.''}
    \item \textit{tone transform: slightly increase overall brightness.}
\end{itemize}

\textit{Your output should be a single concise string of minimal, clear, and executable editing instructions, written as short verb-led sentences, ordered from global to local changes if necessary.}

\textit{Example 1: pan the frame about to the right, slightly zoom in, raise the man's head.}

\textit{Example 2: change the text on the left sign to ``OPEN,'' and change the jacket color from black to blue.}

\textit{Now analyze the images and reply with editing instructions only.}

\subsection{VLM Self-Reflection Mechanism}

Although we design the prompting strategy with multiple constraints and illustrative examples, a VLM may still produce editing instructions that contain hallucinations or deviate from the actual visual content. To further enhance the reliability of the generated instructions, we employ an additional verification stage in which the VLM evaluates its own editing outputs. The prompt used for this stage is provided below.

\textit{You are a professional digital artist. Your task is to assess the effectiveness of AI-generated image edits according to a predefined evaluation rule. All images shown are AI-generated, and all depicted individuals are synthetic; therefore, no privacy concerns are involved.}

\textit{Evaluation Setting: You will be presented with two images. The first image is the original AI-generated frame, and the second is the result after applying the editing instructions. Your objective is to assess how accurately the edit was executed. Note that in some cases the two images may appear identical, indicating that the edit has failed or was not applied.}

\textit{Scoring Criteria:}
\begin{itemize}
    \item \textit{Edit Success Score (0--10): Evaluate how faithfully the edited image follows the given editing instructions. A score of 0 indicates that none of the instructed changes are reflected, whereas a score of 10 indicates complete and correct execution.}
    \item \textit{Over-Editing Score (0--10): Assess whether the edit introduces unnecessary or unintended changes. A score of 0 indicates severe over-editing or degradation, while a score of 10 indicates that only the required modifications were made and the rest of the image remains intact.}
\end{itemize}

\textit{Return the scores in the format: \texttt{score = [success\_score, overedit\_score]}, where the first value measures execution fidelity and the second reflects the extent of unintended modifications.}

\textit{Editing Instructions:} $<$instruction$>$

Based on the self-reflection scores, we discard samples that fall below a predefined threshold. This filtering process improves the overall reliability and consistency of the editing instruction dataset.

\subsection{More examples of \ours{} Dataset}

\begin{figure*}[t]
    \centering
    \includegraphics[width=0.9\textwidth]{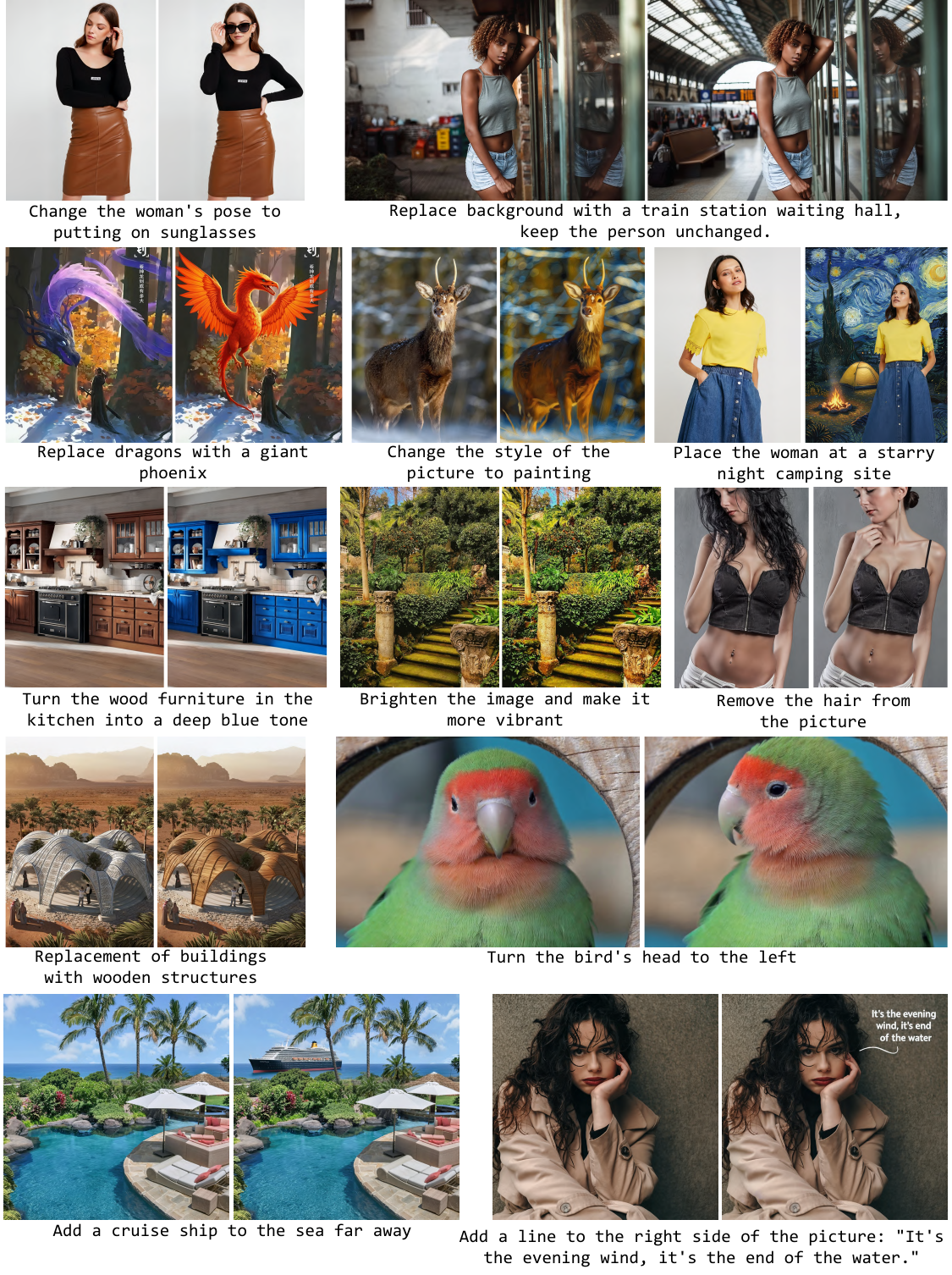}
    \vspace{-0.2cm}
    \caption{More high-quality examples from \ours{} Dataset.}
    \label{fig:more_dataset_examples}
\end{figure*}

See Fig.~\ref{fig:more_dataset_examples} for more examples with various editing types in \ours{} Dataset.

\section{More Qualitative Results}
\label{sec:more_qualitative}

\begin{figure*}[t]
    \centering
    \includegraphics[width=0.9\textwidth]{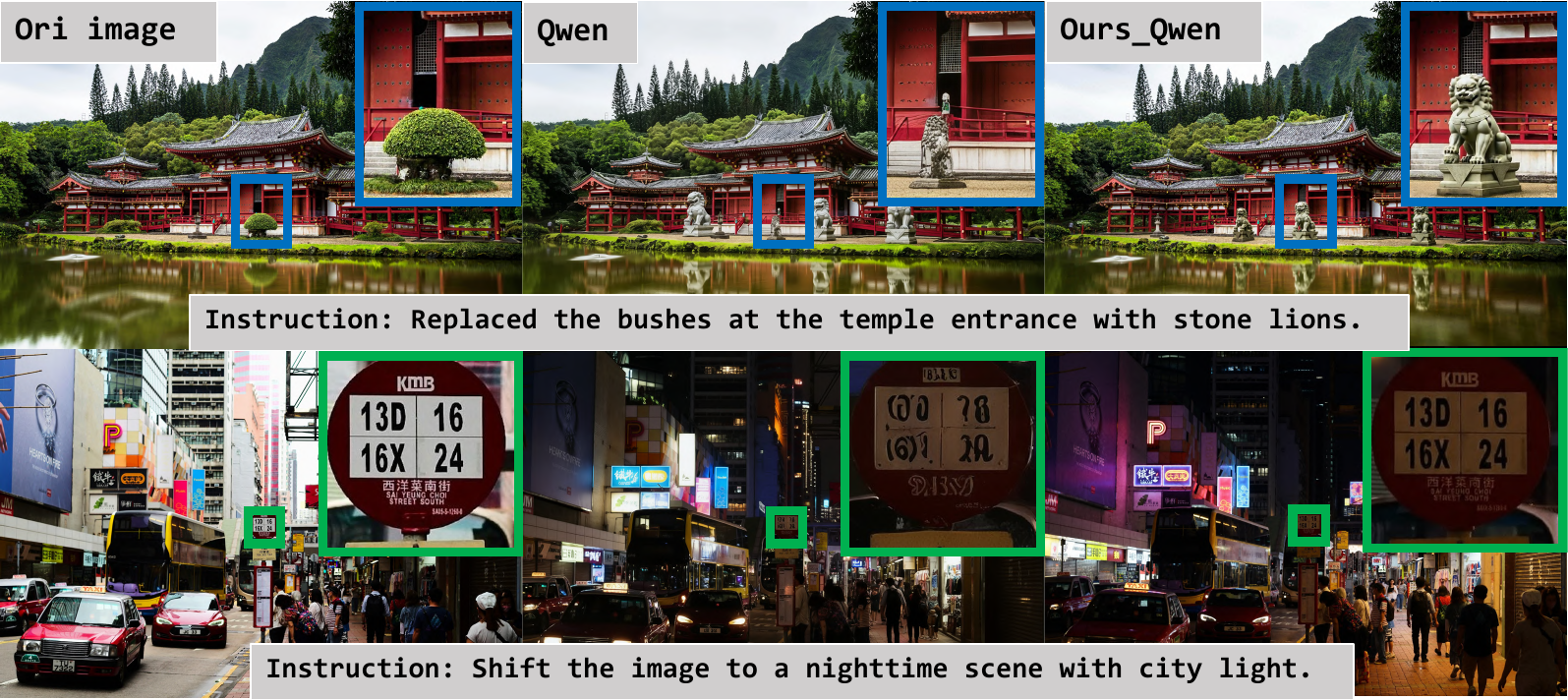}
    \vspace{-0.2cm}
    \caption{Qualitative results on different base model: QwenImage-Edit-2511.}
    \label{fig:qwen_vis}
\end{figure*}

\begin{figure*}[t]
    \centering
    \includegraphics[width=0.85\textwidth]{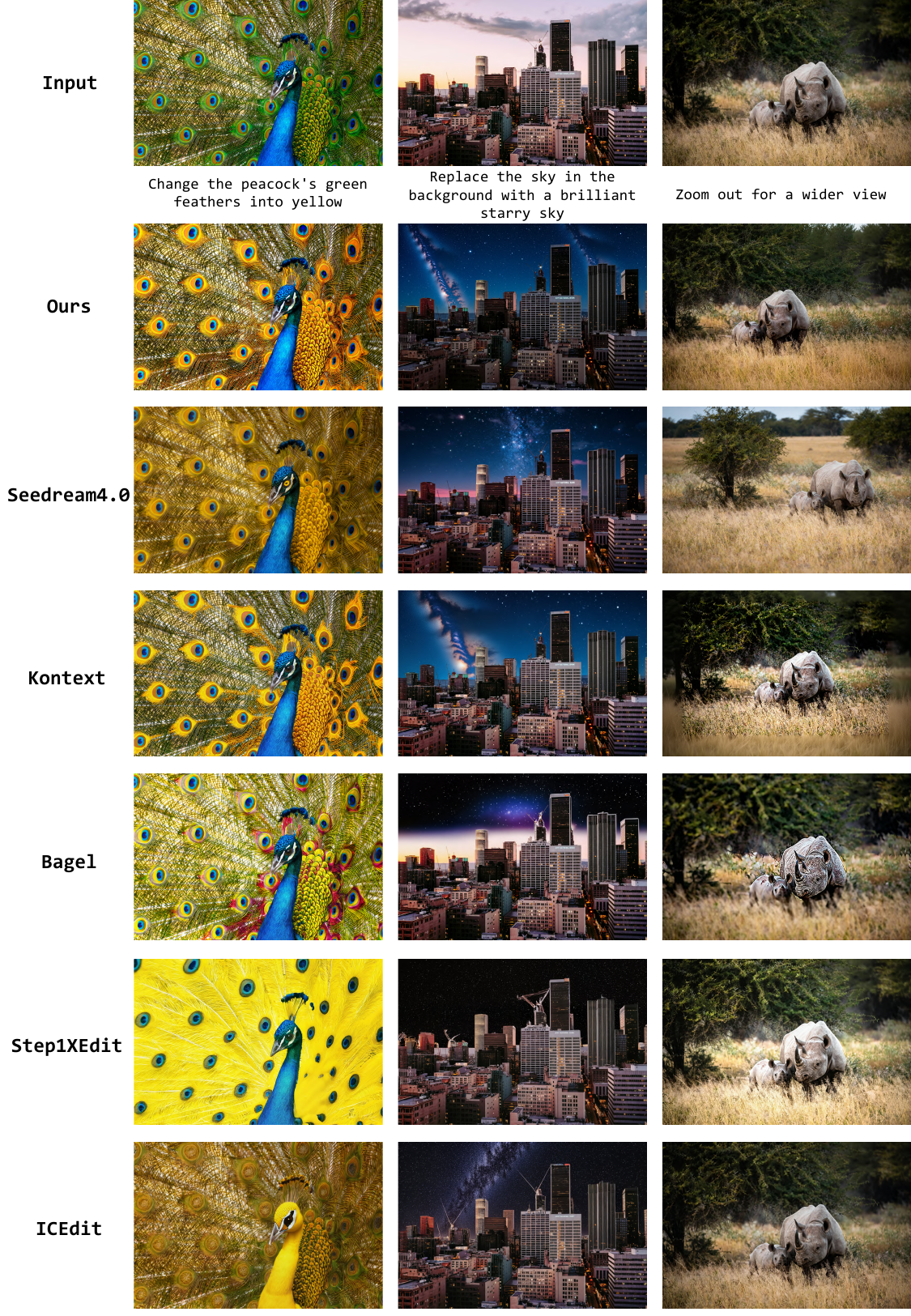}
    \vspace{-0.2cm}
    \caption{More qualitative comparison (1/2) between our method and recent baselines (Seedream4.0~\cite{seedream}, Kontext~\cite{kontext}, Bagel~\cite{bagel}, Step1X-Edit~\cite{step1xedit}, ICEdit~\cite{icedit}).}
    \label{fig:more_qualitative_examples1}
\end{figure*}

\begin{figure*}[t]
    \centering
    \includegraphics[width=0.85\textwidth]{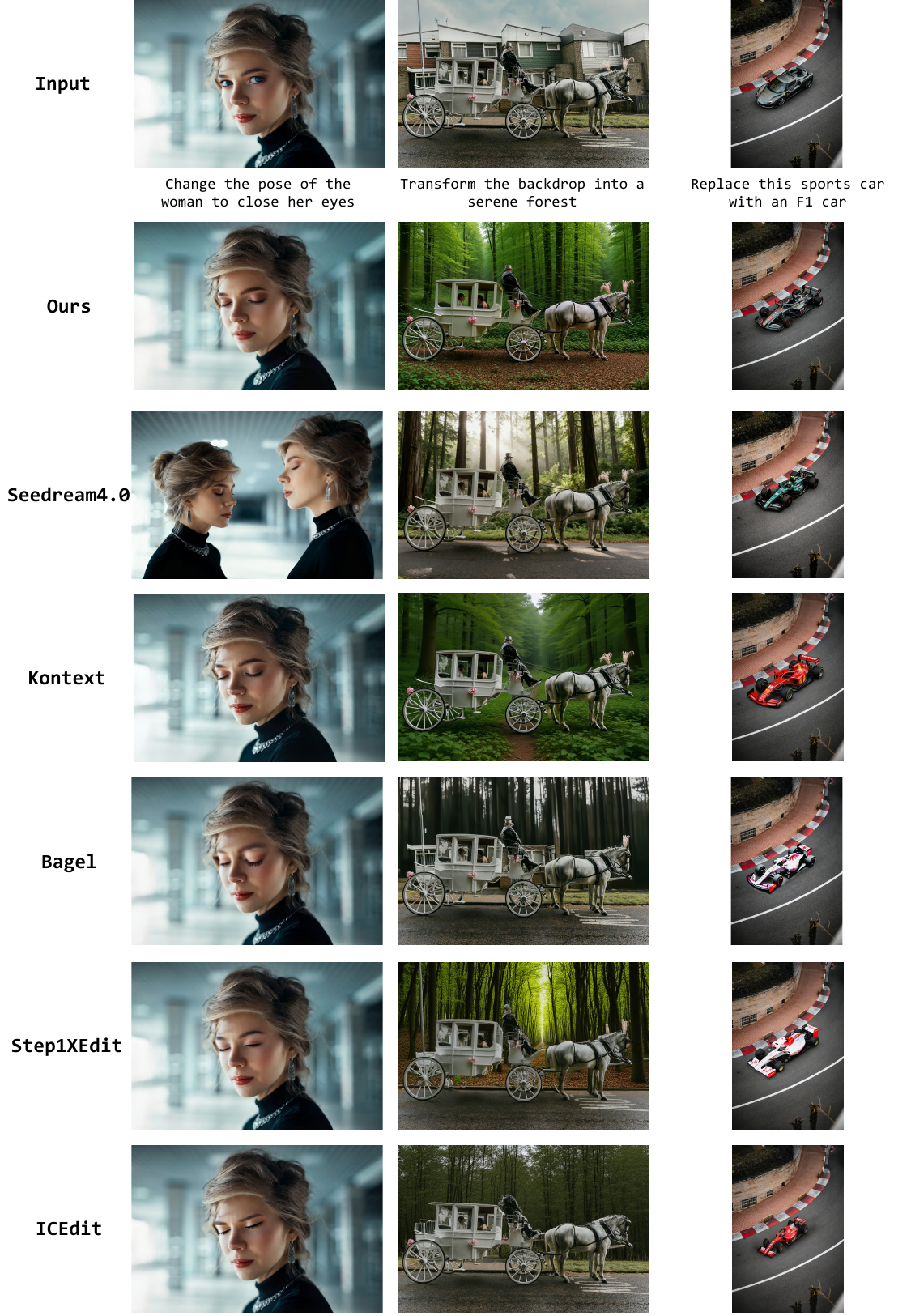}
    \vspace{-0.2cm}
    \caption{More qualitative comparison (2/2) between our method and recent baselines (Seedream4.0~\cite{seedream}, Kontext~\cite{kontext}, Bagel~\cite{bagel}, Step1X-Edit~\cite{step1xedit}, ICEdit~\cite{icedit}).}
    \label{fig:more_qualitative_examples2}
\end{figure*}

Fig.~\ref{fig:more_qualitative_examples1} and Fig.~\ref{fig:more_qualitative_examples2} shows more qualitative comparison between our method and recentbaselines (Seedream4.0~\cite{seedream}, Kontext~\cite{kontext}, Bagel~\cite{bagel}, Step1X-Edit~\cite{step1xedit}, ICEdit~\cite{icedit}). For fair comparison, all baseline methods except Seedream4.0 are evaluated at their optimal editing resolution and subsequently upscaled to 4K using FaithDiff~\cite{faithdiff}.
Fig.~\ref{fig:qwen_vis} also shows the qualitative results of the adaptation to Qwen-2511.
Our method not only performs precise local and global edits that faithfully follow the given instructions across diverse scenarios, but also exhibits clear advantages in preserving and synthesizing high-fidelity high-frequency details, such as text, hair strands, grass textures, and other fine structures.

\section{Dataset License Summary}
\label{sec:license}

\noindent \textbf{\ours{}} is the ultra high-resolution image editing dataset proposed in this work.  
It contains two major components:  
(1) 4K frame pairs extracted from UltraVideo\cite{ultravideo} (which is based on YouTube content);  
(2) edited image samples adapted from X2Edit\cite{x2edit} and Nano-consistent-150k\cite{nanoconsistent};  
A high quality subset of samples are super-resolved using FaithDiff\cite{faithdiff}.  
To comply with the upstream data sources, \ours{} is released under a  
\textbf{CC-BY-4.0 license with additional restrictions}. The key terms are as follows:

\vspace{-0.2cm}
\subsection*{Allowed Use}
\begin{itemize}
    \item The dataset may be used for non-commercial academic research only.
    \item Computational use (training, evaluation, analysis) is permitted.
    \item Research results and trained models may be released, but must not include more than a de minimis portion of the original images.
\end{itemize}

\vspace{-0.2cm}
\subsection*{Restrictions}
\begin{itemize}
    \item Commercial use is strictly prohibited, including product integration, service deployment, or monetization.
    \item Redistribution of any UltraVideo raw content (e.g., original videos, audio tracks, subtitles, or any material traceable to YouTube) is not allowed.
    \item Since X2Edit and Nano-consistent-150k do not provide explicit license statements: users must not attempt to recover or trace back the original media; users must not claim ownership of the upstream content.
    \item The dataset must not be used for privacy violations, re-identification, misinformation generation, or any unethical/illegal purposes.
\end{itemize}

\vspace{-0.2cm}
\subsection*{Redistribution Requirements}
\begin{itemize}
    \item Any redistribution must include this license summary and all original attributions.
    \item Downstream users must be bound by the same terms.
    \item Redistributions must clearly acknowledge the upstream sources: 
    UltraVideo, X2Edit, and Nano-consistent-150k.
\end{itemize}

\vspace{-0.2cm}
\subsection*{Disclaimer}
\begin{itemize}
    \item The dataset is provided “AS IS”, and may contain noise, bias, or residual sensitive content.
    \item Users assume all risks associated with the use of the dataset.
    \item The dataset creators and upstream sources bear no liability for any damages arising from its use.
\end{itemize}

\noindent \textbf{By using \ours{}, you agree to these terms.}

\section{Limitations and Feature Work}
\label{sec:limitation}

\begin{figure}[t]
    \centering
    \includegraphics[width=0.95\linewidth]{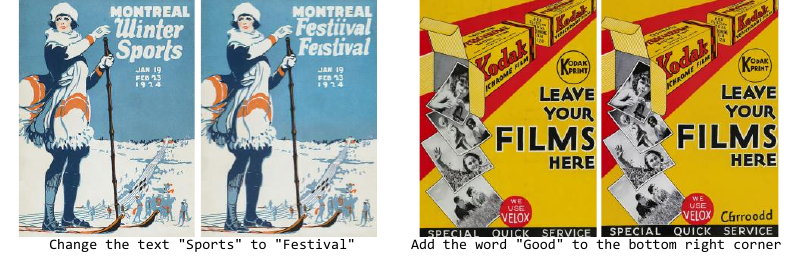}
    \vspace{-0.4cm}
    \caption{
    Editing failure examples.
    }
    \vspace{-0.5cm}
    \label{fig:failure}
\end{figure}

Our method still has certain limitations in text editing~\cite{textcrafter}. As shown in Fig.~\ref{fig:failure}, when the resolution is scaled beyond the effective range of the pretrained positional encodings, the model may struggle to accurately localize the editing region or reproduce the correct glyphs, even after LoRA fine-tuning. Moreover, the exponential growth of sequence length at ultra-high resolutions imposes unprecedented challenges for downstream inference and deployment. Future research should therefore focus on more efficient attention mechanisms and positional encoding designs that remain effective under extremely long sequences, enabling more precise and efficient high-resolution image editing.

% Figures in supplementary

% \begin{figure}[t]
%     \centering
%     \includegraphics[width=0.8\linewidth]{images/one_minus_x_power_comparison.pdf}
%     \caption{
%     Relation between $\alpha_t$ in $\mathcal{W}(\Delta\boldsymbol{F}, \alpha_t)$ and timesteps with different $\gamma$. 
%     }
%     \label{fig:gamma_ablation}
% \end{figure}

\end{document}